%% file: arxiv.tex
\documentclass[letterpaper]{article} 
\usepackage{aaai2027}  
\usepackage[hyphens]{url}  
\usepackage{graphicx} 
\usepackage{natbib}  

\usepackage{caption} 
\usepackage{algorithm}
\usepackage{algorithmic}
\usepackage{amsmath}
\usepackage{amssymb}
\usepackage{multirow}
\usepackage{cuted}
\usepackage{capt-of}
\usepackage{array}
\usepackage[ruled,vlined,linesnumbered,algo2e]{algorithm2e}
\SetKwInput{Input}{input}
\SetKwInput{Output}{output}
\usepackage{cleveref}
\usepackage{newfloat}
\usepackage{listings}
\DeclareCaptionStyle{ruled}{labelfont=normalfont,labelsep=colon,strut=off} 
\floatstyle{ruled}
\newfloat{listing}{tb}{lst}{}
\floatname{listing}{Listing}

\usepackage{booktabs}

\title{DESA-TTA: Dynamic EMA and Source Anchoring for Test-Time Adaptation}
\author{
    Atif Belal, Lilian Hollard, Marco Pedersoli, Eric Granger
}
\affiliations{
    LIVIA, ILLS, Dept. of Systems Engineering, ETS Montréal, Canada\\

    atif.belal.1@ens.etsmtl.ca, \{lilian.hollard, marco.pedersoli, eric.granger\}@etsmtl.ca
}

\begin{document}

\maketitle

\input{0-abstract}
\input{1-introduction}
\input{2-literature_review}
\input{3-method}
\input{4-experiments}
\input{5-conclusion}

\noindent\textbf{Acknowledgments.} This work was supported in part by Distech Controls Inc.,
the Natural Sciences and Engineering Research Council of Canada, the Digital
Research Alliance of Canada, and MITACS.
\input{6-Supplementary}

\makeatletter
\FloatBarrier
\global\@colht\textheight
\global\@colroom\textheight
\global\vsize\textheight
\aaai@orig@bibliography{aaai2027}
\makeatother


%

\end{document}

%% file: 0-abstract.tex
\begin{abstract}
Vision-language object detectors (VLODs) achieve strong zero-shot performance but remain vulnerable to distribution shifts during deployment. Mean-teacher methods for test-time adaptation (TTA) can improve robustness by updating a student model using teacher-generated pseudo-labels. However, mean-teacher TTA is highly sensitive to the choice of a fixed exponential moving average (EMA) coefficient for teacher updates, and repeated optimization with noisy pseudo-labels can cause cumulative student drift.
We propose Dynamic EMA and Source Anchoring for TTA (DESA-TTA), a low-overhead method that jointly regulates teacher updates and student drift through dynamic temporal averaging and source anchoring. Dynamic temporal averaging estimates teacher uncertainty from pseudo-label confidence and box density and uses it to select a sample-wise EMA coefficient within bounds determined by teacher parameter drift. Source anchoring partially restores the updated student parameters toward their pretrained values, with the anchoring strength increasing according to student drift.
Experiments across diverse distribution shifts and two VLOD architectures show consistent improvements over existing TTA methods. On VOC-C, DESA-TTA improves AP$_{50}$ by 14.5 points over zero-shot inference while achieving 55\% higher inference throughput than the previous state-of-the-art TTA method for YOLO-World.\\ 
Our code: \url{https://github.com/imatif17/DESA-TTA}
\end{abstract}

%% file: 1-introduction.tex
\section{Introduction}

VLODs enable open-vocabulary object detection by aligning visual regions with textual class descriptions~\citep{groundingdino,yoloworld}. Unlike conventional closed-set detectors~\citep{fasterrcnn}, VLODs can recognize categories specified at inference time, making them particularly attractive for deployment in dynamic real-world environments. However, their strong zero-shot (ZS) performance can degrade substantially when the deployment distribution differs from the pretraining distribution. To address this limitation, TTA 
has gained increasing attention, as it enables a pretrained model to adapt online during deployment using only unlabeled test data.

State-of-the-art TTA approaches for VLODs primarily follow two paradigms: entropy minimization~\citep{vlodtta} and mean-teacher adaptation~\citep{ttaod-f}. The entropy-minimization approach performs episodic adaptation by resetting the model parameters after each test batch~\citep{vlodtta}. Although this prevents erroneous updates from accumulating, it also limits the model’s ability to retain useful knowledge from previous samples~\citep{sar}. Mean-teacher adaptation instead enables cumulative adaptation by maintaining a temporally averaged teacher that generates pseudo-labels for updating a student model. Given its ability to accumulate target-specific knowledge across the test stream, we focus on mean-teacher adaptation in this work.

\begin{figure*}[t]
    \centering
    \begin{minipage}[t]{0.475\textwidth}
        \centering
        \includegraphics[width=\linewidth]{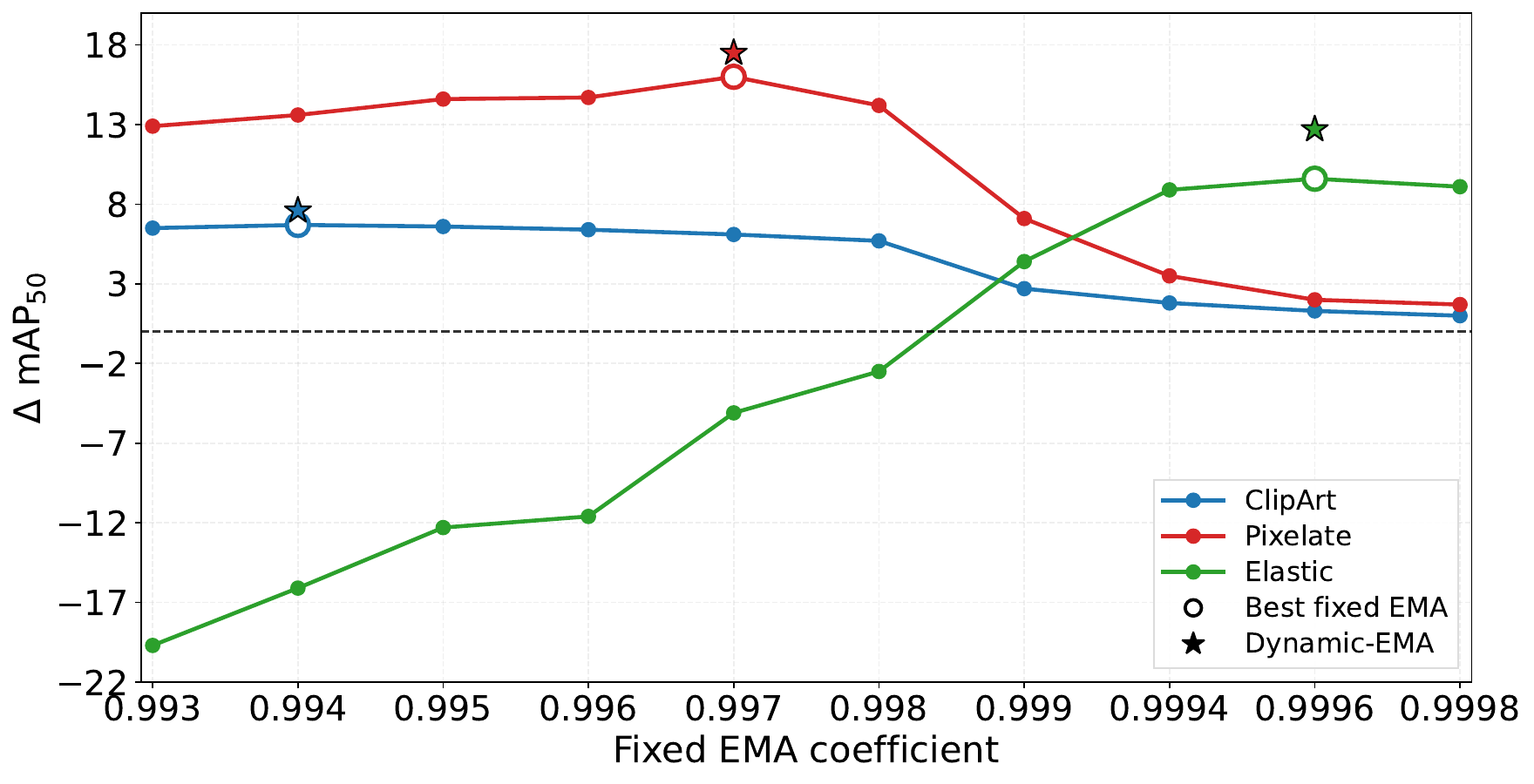}
        \parbox{\linewidth}{\centering\small
        (a) Sensitivity to the fixed EMA coefficient}
    \end{minipage}
    \hfill
    \begin{minipage}[t]{0.475\textwidth}
        \centering
        \includegraphics[width=\linewidth]{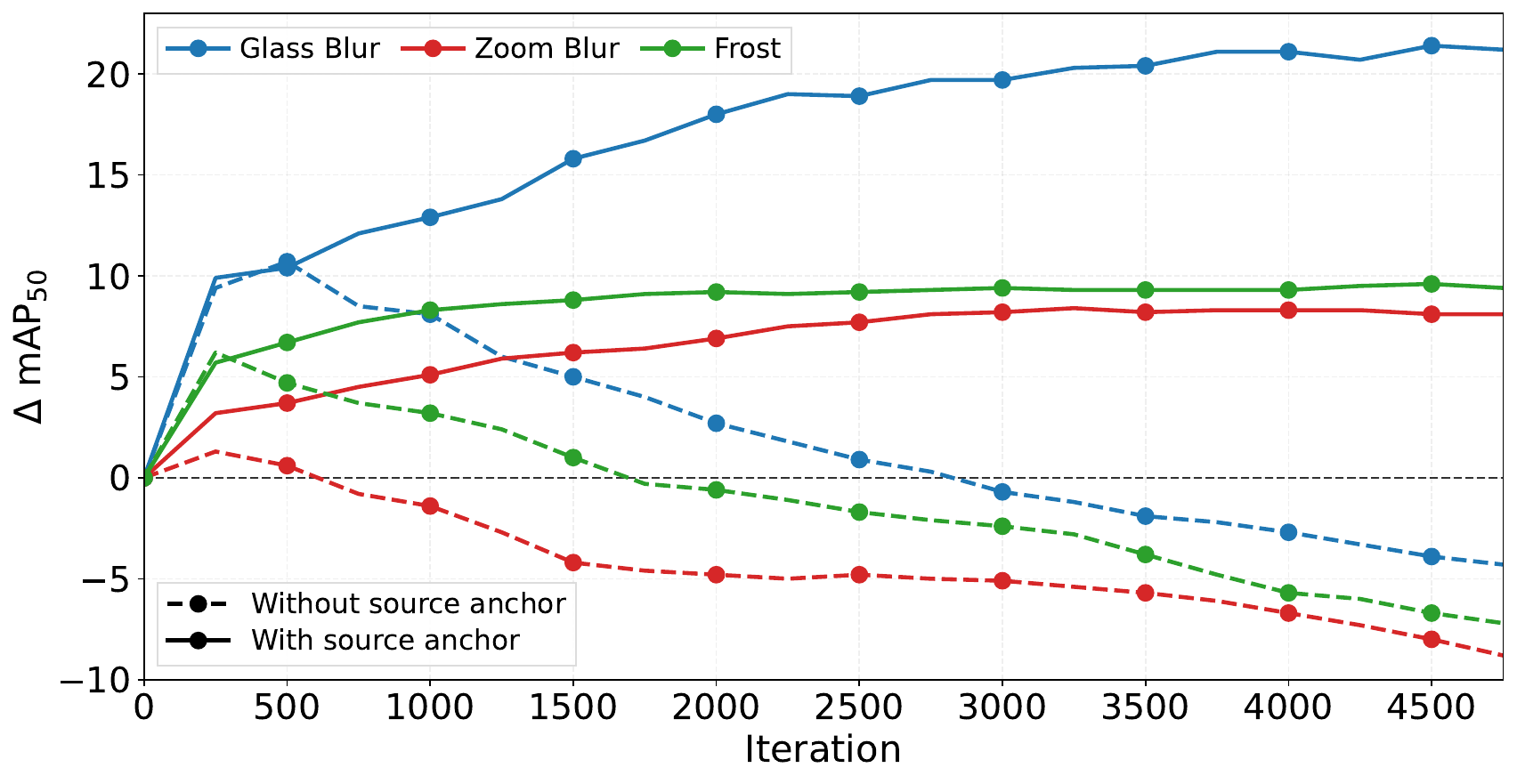}
        \parbox{\linewidth}{\centering\small
        (b) Effect of source anchoring over long streams}
    \end{minipage}
    \vspace{-2mm}
\caption{\textbf{Motivation for DESA-TTA.}
\textbf{(a)} The performance gain from mean-teacher TTA is highly sensitive to the fixed EMA coefficient, whose best-performing value varies substantially across domains. Stars denote dynamic temporal averaging and are placed at the best fixed-EMA coefficient only for visual comparison.
\textbf{(b)} Without source anchoring, the initial adaptation gains diminish as pseudo-label updates accumulate, whereas source anchoring preserves these gains over long target streams.}
    \label{fig:intro_fig}
    \vspace{-4mm}
\end{figure*}

Despite its ability to accumulate knowledge over a test stream, mean-teacher adaptation has two key limitations for TTA of VLODs. First, the teacher relies on a fixed EMA coefficient to balance the preservation of its previous state against the incorporation of the adapted student state. As shown in~\Cref{fig:intro_fig}(a), the preferred coefficient varies substantially across domains. A smaller coefficient incorporates recent student updates more rapidly, enabling faster adaptation but increasing sensitivity to errors induced by noisy pseudo-labels. Conversely, a larger coefficient provides greater stability by preserving the previous teacher state but may make adaptation overly conservative.
Because the target domain is unknown during deployment, selecting a suitable domain-specific coefficient is impractical. Moreover, a fixed coefficient cannot account for variations in pseudo-label reliability or model state over the course of adaptation.

Second, the student is updated cumulatively using pseudo-labels generated throughout the target stream. Since these pseudo-labels can be noisy or miscalibrated, repeated optimization can lead to error accumulation~\citep{cotta}.
As each adapted student state is subsequently incorporated into the teacher, the accumulated errors can affect the pseudo-labels generated for later samples. As shown in~\Cref{fig:intro_fig}(b), adaptation without source anchoring initially improves performance but subsequently deteriorates as more target samples are processed. This behavior indicates that repeated pseudo-label optimization can cause the student to drift progressively away from the pretrained ZS detector, resulting in long-term performance degradation.

To address these limitations, we propose \textbf{DESA-TTA}: \textbf{D}ynamic \textbf{EMA} and \textbf{S}ource \textbf{A}nchoring for TTA of VLODs. DESA-TTA jointly regulates temporal averaging in the teacher and constrains student drift during cumulative adaptation. To reduce sensitivity to a fixed EMA coefficient, DESA-TTA dynamically adjusts the teacher update coefficient according to teacher uncertainty and teacher parameter drift. Teacher uncertainty is estimated from the mean pseudo-label confidence and normalized box density, while teacher drift determines the admissible EMA range. Dense, low-confidence predictions produce greater uncertainty and therefore a smaller coefficient, allowing the teacher to incorporate the adapted student state more rapidly. Conversely, lower uncertainty produces stronger temporal averaging to preserve teacher stability. As shown in~\Cref{fig:intro_fig}(a), dynamic temporal averaging outperforms the best domain-specific fixed-EMA setting on all three displayed domains without requiring prior knowledge of the target distribution.

To mitigate cumulative student drift, DESA-TTA introduces source anchoring after each pseudo-label-based gradient update. The pretrained ZS detector provides a fixed reference, while the adapted student captures target-specific information from the test stream. After each update, the student parameters are partially restored toward their source initialization by linearly interpolating the adapted parameters with their pretrained values using a drift-dependent anchoring coefficient. The anchoring remains weak when the student is close to its initialization and progressively strengthens as student drift increases. This mechanism preserves the student's capacity to acquire target-specific information while constraining excessive deviation from the pretrained detector. As illustrated in~\Cref{fig:intro_fig}(b), source anchoring preserves the initial adaptation gains and substantially mitigates the long-term degradation observed without anchoring. 

The two components address complementary failure modes. Source anchoring constrains the adapted student state before it is incorporated into the teacher, while dynamic temporal averaging controls the extent of this incorporation. Together, they enable cumulative acquisition of target-specific knowledge while limiting error propagation from repeated pseudo-label optimization.

\noindent \textbf{The contributions of this paper are summarized as follows.} 
\textbf{(1)} We identify and empirically characterize two key limitations of mean-teacher TTA for VLODs: the domain sensitivity of fixed EMA updates and cumulative student drift caused by repeated pseudo-label optimization. 
\textbf{(2)} We introduce DESA-TTA, an efficient method that combines sample-wise dynamic temporal averaging with drift-dependent source anchoring to jointly regulate teacher updates and student drift.
\textbf{(3)} Extensive experiments on benchmarks covering style shifts, autonomous-driving scenarios, illumination shifts, and common corruptions show that DESA-TTA consistently outperforms state-of-the-art TTA methods across diverse distribution shifts.

%% file: 2-literature_review.tex
\section{Related Work}

\textbf{Vision-Language Object Detection.  } 
VLODs extend conventional closed-set detectors by aligning visual regions with textual representations, enabling detection beyond a fixed training vocabulary. Early open-vocabulary detection methods leverage image-text pretraining to transfer semantic knowledge from pretrained vision-language models~\citep{clip} to region-level recognition by distilling CLIP representations into detection heads~\citep{vild,regionclip}, expanding the detector vocabulary with image-level supervision~\citep{detic}, or adapting vision-language transformers for open-vocabulary localization~\citep{owlvit}. Grounding-based models unify object detection and phrase grounding by learning region-text alignment from large-scale grounding data~\citep{glip,groundingdino}. More recently, YOLO-World introduces an efficient real-time VLOD by integrating region-text contrastive learning into the YOLO family~\citep{yoloworld}. Despite their strong ZS detection performance, these detectors remain vulnerable to distribution shifts. In this paper, we study the online adaptation of pretrained VLODs using unlabeled target data.

\noindent\textbf{Test-Time Adaptation for Object Detection.  }TTA for object detection (OD-TTA) aims to improve detector robustness by updating the model on unlabeled target images during inference. Prior OD-TTA methods use EMA-based pseudo-label self-training with feature-alignment regularization~\citep{stfar}, IoU-based pseudo-label filtering for fully test-time adaptation~\citep{fullyttaod}, and multi-level feature alignment during online adaptation~\citep{mlfa}. \citet{amrod} further study continual OD-TTA under changing target domains. These methods focus on conventional closed-set detectors and generally assume a shared category space between the source and target domains, limiting their applicability to VLODs.

For VLODs, VLOD-TTA~\citep{vlodtta} introduces a detection-aware entropy objective based on dense proposal overlap but does not explicitly accumulate adaptation over a target stream. TTAOD-F~\citep{ttaod-f} adopts a mean-teacher framework with multi-modal prompt adaptation and instance dynamic memory to improve pseudo-label quality but relies on an additional foundation model, increasing inference cost. In contrast, DESA-TTA enables low-overhead cumulative mean-teacher adaptation by dynamically regulating teacher updates and constraining the adapted model toward its pretrained initialization.

\noindent\textbf{Teacher Stability and Drift Control in TTA. }The mean-teacher framework maintains an EMA teacher whose predictions provide more stable supervision than those of the online student. Prior TTA methods improve this framework through robust consistency learning~\citep{rmt}, timeliness-aware sample reweighting~\citep{rotta}, and parameter-selective teacher updates~\citep{psmt}. However, these methods use a fixed EMA coefficient and therefore cannot adjust the teacher update rate to variations in teacher prediction reliability over the course of adaptation. Moreover, repeated optimization using noisy pseudo-labels can progressively move the student away from its pretrained initialization. To mitigate this drift, CoTTA randomly restores parameters to their source values~\citep{cotta}, \citet{petal} selectively restore parameters according to their estimated importance, \citet{eata} regularize source-important parameters, and \citet{ecotta} constrain adapted features using a frozen source model. These methods are primarily developed for classification and rely on stochastic restoration, parameter-importance estimation, or additional regularization objectives. In contrast, DESA-TTA dynamically adjusts temporal averaging according to teacher uncertainty and parameter drift while mitigating cumulative student drift through deterministic, drift-dependent interpolation toward the pretrained detector.

%% file: 3-method.tex
\begin{figure*}[!t] \centering \includegraphics[width=0.85\textwidth]{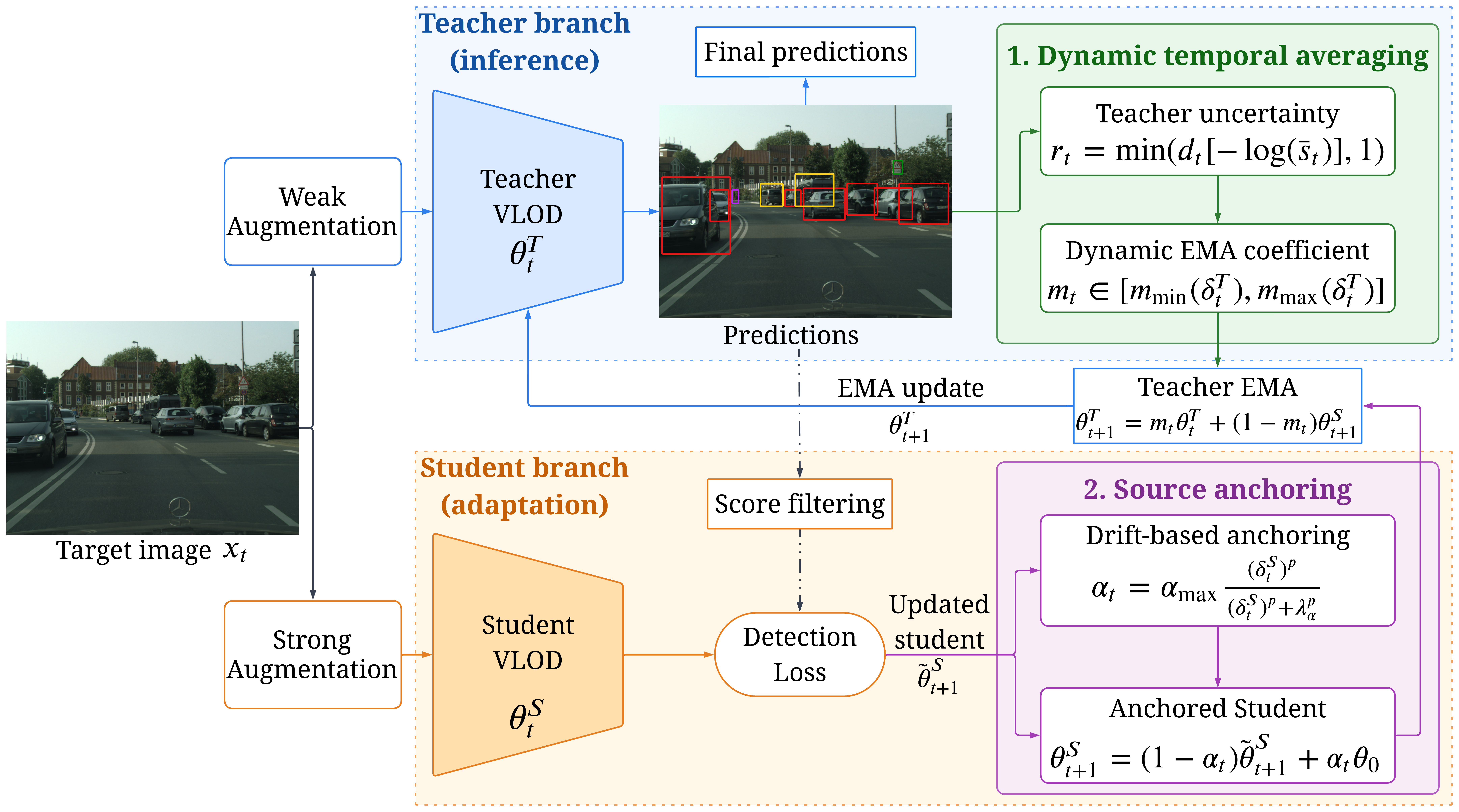} 
\vspace{-2.5mm}
\caption{\textbf{Overview of DESA-TTA.} Given an unlabeled target image, the teacher processes a weakly augmented view to generate pseudo-labels, while the student is optimized using a strongly augmented view. \textbf{(1) Dynamic temporal averaging} computes a teacher-uncertainty score from the mean pseudo-label confidence and box density, and uses it to select $m_t$ within a teacher-drift-dependent EMA range. \textbf{(2) Source anchoring} partially restores the gradient-updated student parameters $\tilde{\theta}^{S}_{t+1}$ toward the pretrained source parameters $\theta_0$ using the drift-dependent coefficient $\alpha_t$, producing the anchored student $\theta^{S}_{t+1}$. The anchored student is subsequently used to update the teacher, whose predictions are used for inference.}
\vspace{-4mm}
\label{fig:desa_overview} 
\end{figure*}

\section{Proposed Methodology}

\subsection{Preliminaries}

\textbf{Vision-Language Object Detector.}
Let $\mathcal{C}=\{c_k\}_{k=1}^{K}$ denote the target vocabulary specified by textual class names. A VLOD comprises an image encoder, a text encoder, and a detection head. Given an input image $x$, the image encoder extracts visual features, and the detection head predicts candidate boxes $\{b_i\}_{i=1}^{N}$. The text encoder maps each class name $c_k$ to a text embedding $t_k$, producing the set $\{t_k\}_{k=1}^{K}$. Each candidate region is represented by a visual embedding $v_i$, with class-wise scores computed as:
\begin{equation}
s_{i,k}=\sigma\left(v_i^\top t_k\right), \quad k=1,\ldots,K ,
\end{equation}
where $\sigma(\cdot)$ is the sigmoid function. After confidence filtering and non-maximum suppression, the detector outputs the final prediction set:
\begin{equation}
\label{eq:detection_prediction}
\mathcal{D}_{\theta}(x,\mathcal{C})=\{(b_j,y_j,s_j)\}_{j=1}^{M},
\end{equation}
where $\theta$ denotes the detector parameters, $M$ is the number of retained detections, and $b_j$, $y_j$, and $s_j$ denote the predicted bounding box, class label, and confidence score of the $j$-th detection, respectively. In this work, a pretrained VLOD is deployed on an unlabeled target stream $\{x_t\}_{t=1}^{T}$ and adapted online without access to source data or target annotations.

\noindent\textbf{Mean-Teacher Test-Time Adaptation.}
At adaptation step $t$, mean-teacher TTA maintains a student model $\mathcal{D}_{\theta_t^S}$ and a teacher model $\mathcal{D}_{\theta_t^T}$, both initialized from the pretrained detector $\mathcal{D}_{\theta_0}$. 
For each target image $x_t$, the teacher generates predictions from a weakly augmented view. After confidence filtering and non-maximum suppression, the retained predictions are used as pseudo-labels to supervise the student on a strongly augmented view through the detection loss, producing the gradient-updated student parameters $\tilde{\theta}_{t+1}^S$. The teacher is subsequently updated using an EMA update: 
\begin{equation}  
\theta_{t+1}^T = m\theta_t^T + (1-m)\tilde{\theta}_{t+1}^S, 
\end{equation}
where $m$ is the EMA coefficient. In standard mean-teacher TTA, $m$ remains fixed, making the teacher update sensitive to target domain characteristics, while repeated pseudo-label 
optimization can induce cumulative student drift.

\subsection{DESA-TTA Method}
An illustration of our DESA-TTA method is shown in~\Cref{fig:desa_overview}. It adapts a pretrained VLOD online through two coupled mechanisms. First, dynamic temporal averaging determines the EMA coefficient used to update the teacher. Second, source anchoring constrains the adapted student state before it is incorporated into the teacher. This joint regulation enables adaptation to the target distribution while reducing sensitivity to the EMA coefficient and limiting cumulative student drift.

\noindent\textbf{Dynamic Temporal Averaging. }
The EMA coefficient $m$ controls the trade-off between preserving the previous teacher state and incorporating the target-adapted student state. However, using a single value of $m$ throughout adaptation is suboptimal because the quality of teacher-generated pseudo-labels can vary substantially across target samples and domains. When the teacher produces a large number of low-confidence predictions, its current state is less reliable for the target sample, and a larger contribution from the target-adapted student may be required. In contrast, when the teacher predictions are confident, stronger temporal averaging is preferable to preserve teacher stability.

For the target image $x_t$, let $M_t$ denote the number of retained teacher pseudo-labels and let $\bar{s}_t$ denote their mean confidence. For $M_t>0$, we define the normalized retained prediction density as $d_t=M_t/M_{\max}$, where $M_{\max}$ is a predefined maximum number of retained detections per image. The teacher-uncertainty score is defined as
\begin{equation}
r_t
=
\min\left(
d_t\left(-\log\bar{s}_t\right),1\right).
\end{equation}
The score increases when the teacher produces a larger number of low-confidence predictions, indicating greater uncertainty in its current predictions. When $M_t=0$, both the student adaptation and the teacher EMA update are skipped.

Although $r_t$ characterizes the uncertainty of the current teacher predictions, it remains a sample-wise signal and does not reflect the extent to which the teacher has changed during adaptation. Consequently, the same uncertainty score may require different levels of temporal averaging depending on the accumulated teacher drift from the pretrained source model. When the drift is small, a lower EMA range facilitates the incorporation of target-specific student updates. As the drift increases, stronger temporal averaging becomes necessary to stabilize the teacher and limit the propagation of accumulated pseudo-label errors. DESA-TTA therefore constrains the dynamic coefficient using a drift-dependent EMA range.

Let $\delta_t^T$ denote the normalized parameter drift of the teacher from the pretrained source model:
\begin{equation}
\delta_t^T = \frac{1}{|\mathcal{A}|}
\sum_{\ell\in\mathcal{A}}
\frac{\left\|\theta_{t,\ell}^T-\theta_{0,\ell}\right\|_1}
{\max\left(\left\|\theta_{0,\ell}\right\|_1,\epsilon\right)},
\end{equation}
where $\mathcal{A}$ denotes the set of adapted parameters and $\epsilon>0$ is a small constant introduced for numerical stability. Based on the teacher drift, DESA-TTA defines separate transition functions for the lower and upper EMA bounds:
\[
g_{\min}(\delta_t^T) = \frac{(\delta_t^T)^p} {(\delta_t^T)^p+\lambda_{\min}^p},
\qquad
g_{\max}(\delta_t^T) = \frac{(\delta_t^T)^p}{(\delta_t^T)^p+\lambda_{\max}^p},
\]
where $\lambda_{\min}$ and $\lambda_{\max}$ determine the drift scales for the lower and upper EMA bounds, respectively, while $p$ controls the sharpness of their transitions.

DESA-TTA defines the lower and upper bounds of the EMA range according to:
\[
\begin{aligned}
m_{\min}(\delta_t^T) &= m_{\min} - \Delta_{\min} \left(1-g_{\min}(\delta_t^T)\right),\\
m_{\max}(\delta_t^T) &= m_{\max} - \Delta_{\max} \left(1-g_{\max}(\delta_t^T)\right),
\end{aligned}
\]
where $m_{\min}$ and $m_{\max}$ denote the asymptotic lower and upper EMA bounds, while $\Delta_{\min}$ and $\Delta_{\max}$ determine their offsets at low teacher drift. When the teacher remains close to the pretrained source model, both bounds take lower values, facilitating the incorporation of target-specific student updates. As $\delta_t^T$ increases, the transition functions approach one and the EMA range gradually shifts toward stronger temporal averaging.

The final dynamic EMA coefficient is selected within this drift-dependent range using the teacher-uncertainty score:
\begin{equation}    
m_t = m_{\max}(\delta_t^T) - \left( m_{\max}(\delta_t^T)-m_{\min}(\delta_t^T) \right)r_t.
\end{equation}

Since $r_t\in[0,1]$, the resulting coefficient remains within $[m_{\min}(\delta_t^T),m_{\max}(\delta_t^T)]$. A larger teacher-uncertainty score therefore produces a smaller EMA coefficient, whereas a lower score results in stronger temporal averaging.

\begin{figure}[t]
\centering
\includegraphics[width=.95\linewidth]{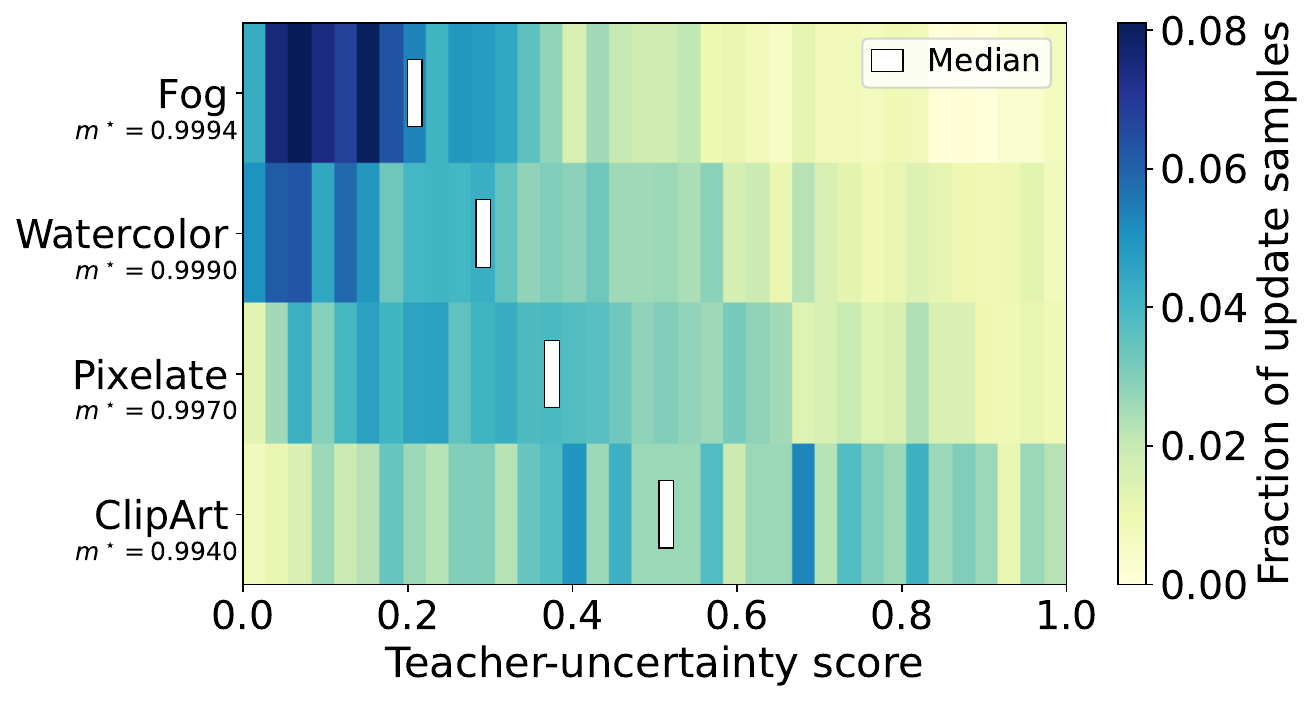}
\vspace{-2.5mm}
\caption{\textbf{Relationship between teacher uncertainty and the preferred EMA coefficient.} The heatmap shows the distribution of the teacher-uncertainty score over update samples. White markers indicate the median score, while $m^\star$ denotes the best fixed EMA coefficient for each domain.}
\vspace{-5mm}
\label{fig:ema_uncertainty}
\end{figure}

As shown in~\Cref{fig:ema_uncertainty}, the preferred fixed EMA coefficient decreases as the median teacher-uncertainty score increases. In particular, Fog has the lowest median uncertainty and favors $m^\star=0.9994$, whereas ClipArt has the highest median uncertainty and favors $m^\star=0.9940$. This observation empirically supports the inverse mapping between $r_t$ and $m_t$. A complementary expected-error analysis motivating this relationship is provided in the supplementary material.

\noindent\textbf{Source Anchoring.}
The student model is updated sequentially using pseudo-labels generated from the target stream. Although these updates are necessary for adaptation, pseudo-label supervision may contain localization and classification errors. Repeated optimization using such pseudo-labels can bias the student trajectory toward accumulated prediction errors, particularly over long target streams. Dynamic temporal averaging regulates the propagation of student updates to the teacher, but it does not directly constrain drift in the student itself. Motivated by the strong ZS capability of the pretrained VLOD, DESA-TTA uses the source initialization as a reference for regularizing the student trajectory while preserving its ability to adapt to the target distribution.

Let $\tilde{\theta}_{t+1,\mathcal{A}}^S$ denote the values of the adapted parameters after the pseudo-label-based gradient update at step $t$. Source anchoring interpolates this adapted state with the corresponding source initialization:
\begin{equation}
\theta_{t+1,\mathcal{A}}^S =
(1-\alpha_t) \tilde{\theta}_{t+1,\mathcal{A}}^S
+
\alpha_t\theta_{0,\mathcal{A}}.
\end{equation}

Because student drift is initially limited but may become more pronounced as pseudo-label errors accumulate over repeated updates, DESA-TTA determines the anchoring strength directly from the current student drift. We define the normalized drift of the gradient-updated student from the source initialization as
\begin{equation}
\delta_t^S = \frac{1}{|\mathcal{A}|}\sum_{\ell\in\mathcal{A}}
\frac{\left\|\tilde{\theta}_{t+1,\ell}^S-\theta_{0,\ell}\right\|_1}
{\max\left(\left\|\theta_{0,\ell}\right\|_1,\epsilon\right)}.
\end{equation}

The anchoring coefficient is then defined as
\begin{equation}
\alpha_t = \alpha_{\max}\frac{(\delta_t^S)^{p}}{(\delta_t^S)^{p}+\lambda_{\alpha}^{p}},
\end{equation}
where $0\leq\alpha_{\max}<1$ denotes the maximum anchoring coefficient, $\lambda_{\alpha}$ determines the anchoring drift scale, and $p$ controls the sharpness of the transition. Consequently, the source constraint remains weak when the student is close to its source initialization and progressively strengthens as student drift increases. Unlike penalty-based $L_2$ regularization, source anchoring constrains the student through post-update parameter interpolation rather than by adding a penalty to the adaptation objective. A proximal interpretation and empirical comparison are provided in the supplementary material.

\noindent\textbf{DESA-TTA Summary.}
Although DESA-TTA can, in principle, adapt any subset of model parameters, in our experiments we set $\mathcal{A}$ to the normalization parameters and keep all remaining parameters fixed. The teacher model is used to generate the final predictions. After source anchoring, the teacher is updated from the anchored student using the dynamic EMA coefficient:
\begin{equation}
\theta_{t+1}^T = m_t\theta_t^T + (1-m_t)\theta_{t+1}^S.
\end{equation}

%% file: 4-experiments.tex
\section{Results and Discussion}

\subsection{Experimental Setup}

\textbf{Datasets. }
We evaluate DESA-TTA under several distribution shifts, including style shifts, autonomous-driving scenarios, low-light conditions, and common corruptions. For style shifts, we use Watercolor, ClipArt, and Comic~\citep{style_shift}. We use Foggy Cityscapes~\cite{foggy_city} and BDD100K~\cite{bdd100k} to evaluate performance under adverse autonomous-driving conditions. ExDark~\cite{exdark} is used to assess performance under low-light conditions. VOC-C and COCO-C~\cite{corruptions} are used to evaluate robustness to 15 common corruption types. These benchmarks cover diverse domain shifts and allow us to assess adaptation behavior over both short and long target streams. Following prior work~\cite{ttaod-f, vlodtta}, each dataset or corruption is treated as an independent single-domain stream. The evaluation therefore considers cumulative online adaptation within a fixed target domain rather than across changing domains.

\noindent\textbf{Baselines. }
Following~\citet{vlodtta}, we compare DESA-TTA with ZS, TPT, VPT, and DPE. The ZS baseline directly evaluates the pretrained VLOD. TPT~\cite{tpt} and VPT~\cite{modprompt} adapt textual and visual prompts, respectively, at test time using target images. DPE~\cite{dpe} adapts dual prototypes at test time to accumulate target-domain knowledge from both visual and textual modalities. We also include a standard mean-teacher (MT) baseline with a fixed EMA coefficient and without source anchoring. Finally, we compare with VLOD-TTA~\cite{vlodtta}, a recent TTA method for VLODs.

\begin{table*}[ht]
\centering
\caption{\textbf{Detection performance on benchmark datasets.} We report mAP and AP$_{50}$ on six benchmark datasets: Watercolor, ClipArt, Comic, Foggy Cityscapes, BDD100K, and ExDark. Best results are in bold.}
\label{tab:main_results}
\resizebox{.8\textwidth}{!}{
\begin{tabular}{lcccccccccccc}
\toprule
\multirow{2}{*}{\textbf{Method}}
& \multicolumn{2}{c}{\textbf{Watercolor}}
& \multicolumn{2}{c}{\textbf{ClipArt}}
& \multicolumn{2}{c}{\textbf{Comic}}
& \multicolumn{2}{c}{\textbf{Foggy Cityscapes}}
& \multicolumn{2}{c}{\textbf{BDD100K}}
& \multicolumn{2}{c}{\textbf{ExDark}} \\
\cmidrule(lr){2-3}
\cmidrule(lr){4-5}
\cmidrule(lr){6-7}
\cmidrule(lr){8-9}
\cmidrule(lr){10-11}
\cmidrule(lr){12-13}
& mAP & AP$_{50}$
& mAP & AP$_{50}$
& mAP & AP$_{50}$
& mAP & AP$_{50}$
& mAP & AP$_{50}$
& mAP & AP$_{50}$ \\
\midrule
ZS       & 26.9 & 47.9 & 24.4 & 40.1 & 17.8 & 29.4 & 13.2 & 21.3 & 13.3 & 22.0 & 35.2 & 64.7 \\
TPT      & 27.3 & 48.5 & 24.9 & 41.3 & 18.1 & 29.9 & 13.3 & 21.5 & 13.4 & 22.2 & 35.8 & 65.1 \\
VPT      & 26.9 & 49.1 & 25.0 & 41.4 & 18.3 & 30.9 & 13.4 & 21.7 & 13.7 & 22.4 & 35.7 & 65.3 \\
DPE      & 27.2 & 48.9 & 24.9 & 41.5 & 18.9 & 31.7 & 13.2 & 21.6 & 13.5 & 22.3 & 35.9 & 65.8 \\
MT       & 29.1 & 50.8 & 27.2 & 45.1 & 21.1 & 36.5 & 13.7 & 22.0 & 12.5 & 20.3 & 35.5 & 65.4 \\
VLOD-TTA & 29.6 & 53.1 & 28.1 & 45.4 & 21.4 & 36.1 & 13.9 & 22.2 & 14.6 & 24.3 & 36.4 & 67.4 \\
\midrule
\textbf{DESA-TTA} & \textbf{31.2} & \textbf{55.0} & \textbf{29.5} & \textbf{48.3} & \textbf{23.0} & \textbf{38.6} & \textbf{15.2} & \textbf{24.4} & \textbf{16.8} & \textbf{27.8} & \textbf{37.3} & \textbf{69.2} \\
\bottomrule
\end{tabular}
}
\vspace{-2mm}
\end{table*}

\begin{table*}[!ht]
\centering
\caption{\textbf{Detection performance on VOC-C.} AP$_{50}$ is reported under 15 common corruptions.}
\label{tab:voc_c_results}
\resizebox{\textwidth}{!}{
\begin{tabular}{lccccccccccccccc|c}
\toprule
\multirow{2}{*}{\textbf{Method}}
& \multicolumn{3}{c}{\textbf{Noise}}
& \multicolumn{4}{c}{\textbf{Blur}}
& \multicolumn{3}{c}{\textbf{Weather}}
& \multicolumn{5}{c}{\textbf{Digital}}
& 
\\
\cmidrule(lr){2-4}
\cmidrule(lr){5-8}
\cmidrule(lr){9-11}
\cmidrule(lr){12-16}
\cmidrule(lr){17-17}
& Gau. & Shot & Imp.
& Def. & Glass & Mot. & Zoom
& Snow & Frost & Fog & Bright.
& Cont. & Elast. & Pix. & JPEG
& Avg. \\
\midrule
ZS       & 16.9 & 17.2 & 16.2 & 32.4 & 11.3 & 26.9 & 30.1 & 46.4 & 50.6 & 70.4 & 73.6 & 41.3 & 42.0 & 8.9  & 26.1 & 34.0 \\
TPT      & 17.4 & 17.9 & 16.4 & 33.1 & 11.7 & 27.1 & 30.7 & 47.2 & 51.9 & 68.9 & 71.1 & 42.8 & 42.7 & 9.5  & 27.2 & 34.4 \\
VPT      & 17.8 & 18.2 & 16.7 & 33.0 & 12.3 & 27.7 & 30.5 & 47.7 & 51.3 & 69.9 & 72.5 & 42.5 & 43.5 & 10.9 & 28.4 & 34.9 \\
DPE      & 17.7 & 18.7 & 17.5 & 32.8 & 12.9 & 28.5 & 30.7 & 48.4 & 52.1 & 70.9 & 73.4 & 42.9 & 43.2 & 11.5 & 30.7 & 35.5 \\
MT       & 27.1 & 28.3 & 28.0 & 33.7 & 14.8 & 28.3 & 30.3 & 51.5 & 54.2 & 69.0 & 70.0 & 55.8 & 51.5 & 10.9 & 39.4 & 39.5 \\
VLOD-TTA & 22.9 & 24.4 & 24.2 & 37.2 & 16.6 & 29.3 & 30.9 & 50.9 & 53.8 & 73.2 & 74.1 & 48.7 & 49.4 & 17.8 & 40.2 & 39.6 \\
\midrule
\textbf{DESA-TTA} & \textbf{34.4} & \textbf{35.8} & \textbf{37.8} & \textbf{39.6} & \textbf{32.7} & \textbf{37.8} & \textbf{36.1} & \textbf{54.5} & \textbf{57.8} & \textbf{73.6} & \textbf{74.2} & \textbf{63.3} & \textbf{59.8} & \textbf{37.9} & \textbf{52.6} & \textbf{48.5} \\
\bottomrule
\end{tabular}
}
\vspace{-4mm}
\end{table*}

\noindent\textbf{Implementation Details. }
We implement DESA-TTA using the pretrained YOLO-World-S checkpoint~\cite{yoloworld}. The student is optimized with AdamW at a learning rate of $6\times10^{-3}$. Following VLOD-TTA, we use a batch size of 1. Teacher pseudo-labels are filtered at a confidence threshold of $0.4$. For dynamic temporal averaging, we set $M_{\max}=300$, $m_{\min}=0.9990$, $\Delta_{\min}=0.0060$, $\lambda_{\min}=0.036$, $m_{\max}=0.9996$, $\Delta_{\max}=0.0016$, and $\lambda_{\max}=0.021$. For source anchoring, we use $\alpha_{\max}=0.035$ and $\lambda_{\alpha}=0.07$. We use $p=2$ as the exponent. These hyperparameters are selected using six corruption types applied to the VOC training split, which contains no images from the evaluation sets, and are then kept fixed across all target domains without target-specific tuning. A detailed hyperparameter sensitivity analysis is provided in the supplementary material. We report AP metrics using the COCO API. Detection results are averaged over three independent runs, with standard deviations provided in the supplementary material. All experiments are run on an RTX 3090 GPU.

\subsection{Comparison with SOTA methods}

\textbf{Style Shift. } 
\Cref{tab:main_results} shows that DESA-TTA achieves the strongest performance across all three style-shift benchmarks. Relative to the ZS baseline, DESA-TTA improves AP$_{50}$ by $+7.1$, $+8.2$, and $+9.2$ on Watercolor, ClipArt, and Comic, respectively. It also outperforms VLOD-TTA by $+1.9$, $+2.9$, and $+2.5$ AP$_{50}$, demonstrating stronger adaptation to appearance changes induced by artistic styles. Compared with standard MT, DESA-TTA further improves AP$_{50}$ by $+4.2$, $+3.2$, and $+2.1$, indicating that a fixed EMA coefficient and unconstrained student adaptation are less effective for cumulative TTA.

\noindent\textbf{Adverse Driving Scenes. }
For adverse driving scenes, DESA-TTA consistently improves performance on both Foggy Cityscapes and BDD100K as shown in~\Cref{tab:main_results}. It increases AP$_{50}$ over the ZS baseline by $+3.1$ on Foggy Cityscapes and $+5.8$ on BDD100K. Compared with VLOD-TTA, DESA-TTA obtains additional gains of $+2.2$ and $+3.5$ AP$_{50}$, respectively. Notably, standard MT degrades performance on BDD100K, which contains a longer target stream and therefore increases the risk of cumulative student drift. In contrast, DESA-TTA remains effective, indicating that source anchoring is important for stable online adaptation in adverse driving scenes.

\noindent\textbf{Illumination Shift. }
For the low-light benchmark, the ExDark results in~\Cref{tab:main_results} show that DESA-TTA obtains the best performance. It improves AP$_{50}$ over the ZS baseline by $+4.5$, showing that online adaptation is beneficial when low-light conditions degrade the reliability of the pretrained detector. DESA-TTA also outperforms VLOD-TTA and standard MT by $+1.8$ and $+3.8$ AP$_{50}$, respectively. These results indicate that dynamic temporal averaging and source anchoring provide more stable adaptation when pseudo-label quality is affected by poor visibility.

\noindent\textbf{Common Corruptions. }
The VOC-C results in~\Cref{tab:voc_c_results} show that DESA-TTA provides strong robustness gains across diverse corruption types. DESA-TTA achieves the highest average AP$_{50}$ across all 15 corruption types, improving from $34.0$ to $48.5$ over the ZS baseline. This corresponds to a gain of $+14.5$ AP$_{50}$. DESA-TTA also outperforms VLOD-TTA and standard MT by $+8.9$ and $+9.0$ AP$_{50}$, respectively. The improvements are particularly pronounced under noise and digital corruptions, where pseudo-label supervision is more error-prone and standard MT can drift. These results show that combining dynamic temporal averaging with source anchoring improves adaptation across diverse corruption types.

\begin{table*}[t]
\centering

\begin{minipage}[t]{0.67\textwidth}
\vspace{0pt}
\centering
\captionof{table}{\textbf{Performance and efficiency comparison using Grounding DINO on COCO-C.} Average mAP for each corruption group is reported along with trainable parameters, GPU memory usage, and latency.}
\label{tab:grounding_dino}

\resizebox{\linewidth}{!}{
\begin{tabular}{lcccc|c|ccc}
\toprule
\multirow{2}{*}{\textbf{Method}} & \multicolumn{5}{c|}{\textbf{Corruption Type Avg. mAP $\uparrow$}} & \multicolumn{3}{c}{\textbf{Efficiency $\downarrow$}} \\
\cmidrule(lr){2-6}
\cmidrule(lr){7-9}
& Noise & Blur & Weather & Digital & Avg. & \shortstack{Trainable\\Params. (M)} & \shortstack{Memory\\(GB)} & \shortstack{Latency\\(ms/img)} \\
\midrule
ZS & 14.9 & 11.2 & 34.7 & 23.0 & 20.6 & 0.00 & 1.36 & 210.6 \\
TTAOD-F & 21.2 & 14.3 & 37.0 & 31.7 & 26.0 & \textbf{0.08} & 11.37 & 701.9 \\
VLOD-TTA & 21.1 & 15.6 & 38.5 & 30.4 & 26.2 & 0.89 & 3.76 & 531.6 \\
\midrule
\textbf{DESA-TTA} & \textbf{22.6} & \textbf{15.9} & \textbf{38.9} & \textbf{33.1} & \textbf{27.6} & 0.09 & \textbf{3.37} & \textbf{492.5} \\
\bottomrule
\end{tabular}
}
\end{minipage}
\hfill
\begin{minipage}[t]{0.31\textwidth}
\vspace{1.5pt}
\centering
\captionof{table}{\textbf{Accuracy and efficiency comparison on VOC-C.} FPS, trainable parameters, and average AP$_{50}$ are reported.}
\label{tab:efficiency}
\resizebox{\linewidth}{!}{
\begin{tabular}{lccc}
\toprule
\textbf{Method} & \textbf{FPS $\uparrow$} & \textbf{Params (M) $\downarrow$} & \textbf{AP$_{50}$ $\uparrow$} \\
\midrule
ZS       & 89 & 0    & 34.0 \\
TPT      & 9  & 1.12 & 34.4 \\
VPT      & 18 & 3.93 & 34.9 \\
DPE      & 15 & 0.31 & 35.5 \\
VLOD-TTA & 20 & 1.61 & 39.6 \\
\midrule
\textbf{DESA-TTA} & \textbf{31} & \textbf{0.021} & \textbf{48.5} \\
\bottomrule
\end{tabular}
}
\end{minipage}

\vspace{-4mm}
\end{table*}

\subsection{Ablation Studies}

\noindent\textbf{Contribution of Individual Components. }
Table~\ref{tab:component_ablation} shows the contribution of each DESA-TTA component. Starting from MT, the teacher-drift-dependent EMA range provides limited gains, especially on Style, indicating that adjusting the EMA range based on teacher drift is insufficient by itself.
Using the teacher-uncertainty score without the drift-dependent range achieves larger gains on Style and comparable gains on VOC-C, reaching $45.6$ and $41.4$ AP$_{50}$, respectively. This shows that the sample-wise uncertainty signal is informative but remains insufficient to achieve the best performance. Among the two uncertainty cues, confidence is more effective than box density alone, while combining both cues yields the strongest dynamic EMA (D-EMA) result. Full D-EMA improves AP$_{50}$ to $46.6$ on Style and $43.1$ on VOC-C, confirming that the teacher-drift-dependent range and uncertainty cues are complementary.
For source anchoring, adding a static source anchor to MT improves performance over standard MT, while student-drift-dependent anchoring further improves both benchmarks. Therefore, the anchoring coefficient should adapt to current student drift rather than remain fixed. Finally, combining D-EMA with student-drift-dependent source anchoring yields the best results, reaching $47.3$ AP$_{50}$ on Style and $48.5$ AP$_{50}$ on VOC-C. These results indicate that the two mechanisms address complementary failure modes: source anchoring constrains the adapted student state before the teacher update, while dynamic temporal averaging regulates its incorporation into the teacher.

\begin{table}[t]
\centering
\caption{\textbf{Component ablation of DESA-TTA.} Average AP$_{50}$ is reported over the three style-shift datasets and the 15 VOC-C corruptions. Here, $\delta_t^T$ indicates the use of teacher drift to determine the EMA range, $r_t$ denotes the teacher-uncertainty score, and $\bar{s}_t$ and $d_t$ denote the confidence and density cues used to compute $r_t$. SA denotes static source anchoring, D-SA denotes source anchoring controlled by $\alpha_t$, and $\alpha_t$ is the student-drift-dependent anchoring coefficient.}
\label{tab:component_ablation}
\setlength{\tabcolsep}{4.5pt}
\resizebox{.45\textwidth}{!}{
\begin{tabular}{lcccccc|cc}
\toprule
\multirow{2}{*}{Method}
& \multicolumn{4}{c}{D-EMA}
& \multicolumn{2}{c}{SA}
& \multicolumn{2}{c}{AP$_{50}$} \\
\cmidrule(lr){2-5}
\cmidrule(lr){6-7}
\cmidrule(lr){8-9}
& $\delta_t^T$ & $r_t$ & $\bar{s}_t$ & $d_t$ & SA & $\alpha_t$ & Style & VOC-C \\
\midrule
ZS        & -- & -- & -- & -- & -- & -- & 39.1 & 34.0 \\
MT        & -- & -- & -- & -- & -- & -- & 44.1 & 39.5 \\
\midrule
$\delta_t^T$ only      & \checkmark & --         & --         & --         & --         & --         & 44.2 & 41.2 \\
$r_t$ only      & --         & \checkmark & \checkmark & \checkmark & --         & --         & 45.6 & 41.4 \\
$\bar{s}_t$ only & \checkmark & \checkmark & \checkmark & --         & --         & --         & 46.1 & 42.7 \\
$d_t$ only       & \checkmark & \checkmark & --         & \checkmark & --         & --         & 43.8 & 41.3 \\
D-EMA            & \checkmark & \checkmark & \checkmark & \checkmark & --         & --         & 46.6 & 43.1 \\
\midrule
MT+SA     & --         & --         & --         & --         & \checkmark & --         & 45.7 & 40.6 \\
MT+D-SA   & --         & --         & --         & --         & \checkmark & \checkmark & 46.3 & 42.2 \\
\midrule
D-EMA+SA  & \checkmark & \checkmark & \checkmark & \checkmark & \checkmark & --         & 46.4 & 45.8 \\
\textbf{DESA-TTA} & \checkmark & \checkmark & \checkmark & \checkmark & \checkmark & \checkmark & \textbf{47.3} & \textbf{48.5} \\
\bottomrule
\end{tabular}
}
\vspace{-5mm}
\end{table}

\noindent\textbf{Extension to Grounding DINO.}
To evaluate the generalizability of DESA-TTA across VLOD architectures, we extend it to Grounding DINO \cite{groundingdino} and evaluate it on COCO-C. This also enables a direct comparison with TTAOD-F~\citep{ttaod-f}, another mean-teacher-based method whose applicability is limited to transformer-based architectures. As shown in~\Cref{tab:grounding_dino}, DESA-TTA consistently outperforms the ZS model and both TTA baselines across all corruption groups. It improves the average mAP from 20.6 to 27.6, surpassing TTAOD-F and VLOD-TTA by 1.6 and 1.4 points, respectively.
DESA-TTA also provides a favorable accuracy--efficiency trade-off. It adapts only 0.09M parameters, compared with 0.89M for VLOD-TTA. Moreover, DESA-TTA requires 3.37\,GB of GPU memory and 492.5\,ms per image, reducing memory consumption and latency by 10.4\% and 7.4\%, respectively, compared with VLOD-TTA. Compared with TTAOD-F, it reduces memory consumption by 70.4\% and latency by 29.8\%. These results demonstrate that DESA-TTA remains effective across both CNN- and transformer-based VLOD architectures.

\noindent\textbf{Runtime Cost. }
TTA methods adapt the model during inference, introducing additional latency compared with ZS evaluation. Table~\ref{tab:efficiency} compares the efficiency of DESA-TTA with existing TTA baselines on YOLO-World. Compared with TPT, VPT, DPE, and VLOD-TTA, DESA-TTA adapts substantially fewer parameters and achieves higher inference speed than all adaptation baselines. At the same time, it obtains the highest AP$_{50}$ of $48.5$ on VOC-C, demonstrating a favorable efficiency--performance trade-off among TTA methods. The supplementary material further shows that, for a stationary target domain, DESA-TTA can be adapted once on an unlabeled target stream and subsequently deployed at full inference speed.

%% file: 5-conclusion.tex
\section{Conclusion}
We studied cumulative mean-teacher TTA for VLODs and identified two key limitations: the domain sensitivity of a fixed EMA coefficient and cumulative student drift caused by repeated pseudo-label optimization. To address these limitations, we introduced DESA-TTA, combining dynamic temporal averaging with source anchoring. 
Dynamic temporal averaging regulates teacher updates according to teacher uncertainty and teacher drift, while source anchoring constrains the adapted student toward its pretrained initialization according to student drift. Experiments under style shifts, adverse driving conditions, low-light conditions, and common corruptions show consistent improvements across YOLO-World and Grounding DINO. Ablation studies further show that the two components address complementary failure modes and achieve the strongest performance when combined. Overall, DESA-TTA provides a cost-effective and stable method for adapting VLODs to unlabeled target streams without access to source data.\\
\textbf{Supplementary material.} It provides theoretical analysis, implementation details, additional results on backbone generalization, random-seed variation, and held-out target data, design ablations, comparisons with source-anchoring methods, hyperparameter analysis, and qualitative results.


%% file: 6-Supplementary.tex
\section{Appendix}

\subsection{Additional Theoretical Analysis}

\noindent\textbf{Expected-Error Analysis of Dynamic Temporal Averaging. }

\noindent\textbf{Proposition 1.}
Let $\theta_t^\star$ denote an unknown target-optimal parameter state, and let
\[
e_t^T=\theta_t^T-\theta_t^\star,
\qquad
e_t^S=\theta_{t+1}^S-\theta_t^\star
\]
denote the errors of the current teacher and the student state used in the EMA update, respectively. Conditioned on a teacher-uncertainty level $r$, define
\[
V_T(r) = \mathbb{E} \left[ \|e_t^T\|_2^2 \mid r_t=r \right],
\]
\[
V_S(r) = \mathbb{E} \left[ \|e_t^S\|_2^2 \mid r_t=r \right],
\]
and
\[
C(r) = \mathbb{E} \left[(e_t^T)^\top e_t^S \mid r_t=r \right]. \]

For an EMA update
\[
\theta_{t+1}^T = m\theta_t^T + (1-m)\theta_{t+1}^S,
\]
when the teacher and student errors are not identical almost surely, the unique unconstrained coefficient minimizing the expected squared parameter error is
\[
m^\star(r) = \frac{V_S(r)-C(r)} {V_T(r)+V_S(r)-2C(r)}.
\]
Furthermore, for fixed $V_S(r)$ and $C(r)$, if $V_S(r)>C(r)$, then the optimal unconstrained EMA coefficient decreases with $V_T(r)$:
\[
\frac{\partial m^\star}{\partial V_T}<0.
\]

\noindent\textit{Proof.}
The error of the EMA-updated teacher relative to $\theta_t^\star$ is
\[
e_{t+1}^T = \theta_{t+1}^T-\theta_t^\star = m e_t^T+(1-m)e_t^S.
\]
Therefore, conditioned on $r_t=r$, its expected squared error is
\[
\begin{aligned}
\mathcal{R}(m\mid r)
&= \mathbb{E} \left[ \left\| m e_t^T+(1-m)e_t^S \right\|_2^2 \mid r_t=r \right] \\
&= m^2V_T(r) + (1-m)^2V_S(r) + 2m(1-m)C(r).
\end{aligned}
\]
Differentiating with respect to $m$ gives
\[
\frac{\partial\mathcal{R}(m\mid r)}{\partial m} = 2m \left( V_T(r)+V_S(r)-2C(r) \right) - 2\left(V_S(r)-C(r) \right).
\]
Setting the derivative to zero yields
\[
m^\star(r) = \frac{V_S(r)-C(r)}{V_T(r)+V_S(r)-2C(r)}.
\]
Moreover,
\[
\frac{\partial^2\mathcal{R}(m\mid r)}{\partial m^2} = 2 \left(V_T(r)+V_S(r)-2C(r)\right).
\]
Since
\[
V_T(r)+V_S(r)-2C(r) = \mathbb{E} \left[\|e_t^T-e_t^S\|_2^2\mid r_t=r \right] > 0,
\]
the objective is strictly convex in $m$, and the stationary point is its unique unconstrained minimizer. If the EMA coefficient is constrained to $[0,1]$, the solution is the projection of the unconstrained minimizer onto this interval.

For fixed $V_S(r)$ and $C(r)$, differentiating $m^\star(r)$ with respect to $V_T(r)$ gives
\[
\frac{\partial m^\star}{\partial V_T} = - \frac{V_S(r)-C(r)} {\left(V_T(r)+V_S(r)-2C(r)\right)^2}.
\]
Thus, when $V_S(r)>C(r)$,
\[
\frac{\partial m^\star}{\partial V_T}<0.
\]
Therefore, when a larger teacher-uncertainty score reflects greater teacher error, assigning a smaller EMA coefficient is consistent with reducing the expected error of the EMA update. This result establishes the decreasing direction of the uncertainty--EMA relationship; it does not claim that the specific bounded linear mapping used by DESA-TTA is the exact optimal mapping.\\

\noindent\textbf{Proximal Interpretation and Contraction of Source Anchoring. }

\noindent\textbf{Proposition 2.}
Let $\tilde{\theta}_{t+1,\mathcal{A}}^S$ denote the adapted parameter subset after the pseudo-label-based gradient update, and let $\theta_{0,\mathcal{A}}$ denote its source initialization. For $\lambda_t\geq0$, consider
\[
\theta_{t+1,\mathcal{A}}^S = \arg\min_{\theta_{\mathcal{A}}} \frac{1}{2} \left\| \theta_{\mathcal{A}} -\tilde{\theta}_{t+1,\mathcal{A}}^S \right\|_2^2
+
\frac{\lambda_t}{2} \left\| \theta_{\mathcal{A}} - \theta_{0,\mathcal{A}} \right\|_2^2.
\]
Its unique solution is
\[
\theta_{t+1,\mathcal{A}}^S = (1-\alpha_t) \tilde{\theta}_{t+1,\mathcal{A}}^S + \alpha_t\theta_{0,\mathcal{A}}, \qquad
\alpha_t = \frac{\lambda_t}{1+\lambda_t}.
\]
Furthermore, this update scales the post-gradient distance from the source initialization by exactly the factor $1-\alpha_t$:
\[
\left\| \theta_{t+1,\mathcal{A}}^S - \theta_{0,\mathcal{A}} \right\|_1 = (1-\alpha_t) \left\| \tilde{\theta}_{t+1,\mathcal{A}}^S - \theta_{0,\mathcal{A}} \right\|_1.
\]

\noindent\textit{Proof.}
Define
\[
\mathcal{J}(\theta_{\mathcal{A}}) = \frac{1}{2} \left\| \theta_{\mathcal{A}} - \tilde{\theta}_{t+1,\mathcal{A}}^S \right\|_2^2
+
\frac{\lambda_t}{2} \left\| \theta_{\mathcal{A}} - \theta_{0,\mathcal{A}} \right\|_2^2.
\]
Its gradient is
\[
\nabla_{\theta_{\mathcal{A}}}\mathcal{J} = \theta_{\mathcal{A}} - \tilde{\theta}_{t+1,\mathcal{A}}^S 
+
\lambda_t \left( \theta_{\mathcal{A}} - \theta_{0,\mathcal{A}} \right).
\]
Setting the gradient to zero gives
\[
(1+\lambda_t)\theta_{\mathcal{A}} = \tilde{\theta}_{t+1,\mathcal{A}}^S + \lambda_t\theta_{0,\mathcal{A}}.
\]
Therefore,
\[
\theta_{t+1,\mathcal{A}}^S = \frac{1}{1+\lambda_t} \tilde{\theta}_{t+1,\mathcal{A}}^S + \frac{\lambda_t}{1+\lambda_t} \theta_{0,\mathcal{A}}.
\]
Because
\[
\alpha_t = \frac{\lambda_t}{1+\lambda_t},  \qquad
1-\alpha_t = \frac{1}{1+\lambda_t},
\]
this is exactly the source-anchoring update
\[
\theta_{t+1,\mathcal{A}}^S = (1-\alpha_t) \tilde{\theta}_{t+1,\mathcal{A}}^S + \alpha_t\theta_{0,\mathcal{A}}.
\]
The Hessian of $\mathcal{J}$ is
\[
\nabla^2_{\theta_{\mathcal{A}}}\mathcal{J} = (1+\lambda_t)I,
\]
which is positive definite for $\lambda_t\geq0$. Hence, the solution is unique.

Subtracting $\theta_{0,\mathcal{A}}$ from the anchoring update gives
\[
\begin{aligned}
\theta_{t+1,\mathcal{A}}^S - \theta_{0,\mathcal{A}} 
&= (1-\alpha_t) \tilde{\theta}_{t+1,\mathcal{A}}^S + \alpha_t\theta_{0,\mathcal{A}} - \theta_{0,\mathcal{A}} \\
&= (1-\alpha_t) \left(\tilde{\theta}_{t+1,\mathcal{A}}^S - \theta_{0,\mathcal{A}} \right).
\end{aligned}
\]
Since $0\leq\alpha_t<1$, taking the $\ell_1$ norm yields
\[
\left\| \theta_{t+1,\mathcal{A}}^S - \theta_{0,\mathcal{A}} \right\|_1
= (1-\alpha_t) \left\| \tilde{\theta}_{t+1,\mathcal{A}}^S - \theta_{0,\mathcal{A}} \right\|_1.
\]
Thus, source anchoring retains a $1-\alpha_t$ fraction of the post-gradient parameter distance from the source initialization, equivalently removing an $\alpha_t$ fraction of that distance.

\subsection{Implementation and Evaluation Protocol}

\noindent\textbf{Online Adaptation Procedure. }
Algorithm~\ref{alg:desa_tta} summarizes the online adaptation procedure of DESA-TTA. For each incoming target image, the current teacher first produces the detection output used for evaluation. The same teacher detections are then filtered and used as pseudo-labels to adapt the student. After the student update, source anchoring partially restores the adapted parameters toward the pretrained model, and the teacher is updated with the dynamic EMA coefficient.\\

\begin{algorithm2e}[t]
\DontPrintSemicolon
\caption{DESA-TTA Online Adaptation}
\label{alg:desa_tta}

\Input{pretrained VLOD $\theta_0$, target stream $\{x_t\}_{t=1}^{T}$, class names $\mathcal{C}$}
\Output{predictions for the target stream}

Initialize student model $\theta^S \leftarrow \theta_0$\;
Initialize teacher model $\theta^T \leftarrow \theta_0$\;

\For{$t \leftarrow 1$ \KwTo $T$}{
    Predict with the current teacher:
    $\mathcal{P}_t \leftarrow D_{\theta^T}(x_t,\mathcal{C})$\;
    
    Use $\mathcal{P}_t$ as the output prediction for image $x_t$\;

    Filter $\mathcal{P}_t$ to obtain pseudo-labels $\hat{\mathcal{Y}}_t$\;

    \eIf{$\hat{\mathcal{Y}}_t$ is empty}{
        Keep the models unchanged\;
    }{
        Compute teacher uncertainty from pseudo-label confidence and box density\;

        Compute dynamic EMA coefficient $m_t$\;

        Update student $\theta^S$ using pseudo-labels $\hat{\mathcal{Y}}_t$\;

        Compute source anchoring coefficient $\alpha_t$\;

        Partially restore $\theta^S$ toward $\theta_0$\;

        Update teacher for the next target image:
        $\theta^T \leftarrow m_t\theta^T + (1-m_t)\theta^S$\;
    }
}

\end{algorithm2e}

\noindent\textbf{Additional Implementation Details. }
For each target stream, the student and teacher are initialized from the same pretrained VLOD checkpoint and target images are processed sequentially with batch size one. Each image is used once, and no source images or target annotations are used during adaptation. The teacher branch uses a weak view of the target image for detection and pseudo-label generation. This weak view applies YOLOv5 keep-ratio resizing to $640\times640$, letterbox resizing with padding value 114, text loading, and input packing.

The student branch uses a strong view of the same image for pseudo-label supervision. This strong view applies YOLOv5 keep-ratio resizing to 640×640, random horizontal flipping with probability 0.5, photometric distortion, random erasing with probability 0.7, letterbox resizing with padding value 114, text loading, and input packing. Teacher detections are filtered by confidence thresholding and NMS before being used as pseudo-labels. If no pseudo-label is retained, the optimizer step, source anchoring step, and teacher EMA update are skipped for that image.

For YOLO-World experiments, only the batch-normalization parameters are adapted, while all other parameters are kept fixed. The student is optimized with AdamW using a learning rate of $6\times10^{-3}$, and one gradient update is performed per target image. For Grounding DINO, we adapt the layer-normalization and group-normalization parameters while keeping the remaining parameters fixed. After each student update, source anchoring is applied to the adapted parameters, and the teacher is updated using the dynamic EMA coefficient. Each dataset or corruption type is adapted independently from the pretrained initialization.\\

\noindent\textbf{Dataset Details. }
Table~\ref{tab:dataset_details} summarizes the target datasets used in our experiments. For VOC-C and COCO-C, each corruption type is processed as an independent target stream.

\begin{table}[t]
\centering
\caption{\textbf{Summary of target datasets used in the experiments.}}
\label{tab:dataset_details}
\resizebox{\linewidth}{!}{
\begin{tabular}{lccc}
\toprule
Dataset & Shift Type & \# Target Images & \# Classes \\
\midrule
Watercolor & Artistic style shift & 1,000 & 6 \\
ClipArt & Artistic style shift & 500 & 20 \\
Comic & Artistic style shift & 1,000 & 6 \\
Foggy Cityscapes & Synthetic fog & 500 & 8 \\
BDD100K & Real-world driving scenes & 10,000 & 10 \\
ExDark & Low-light scenes & 7,363 & 12 \\
VOC-C & Common corruptions & 4,952 per corruption & 20 \\
COCO-C & Common corruptions & 5,000 per corruption & 80 \\
\bottomrule
\end{tabular}
}
\end{table}

\subsection{Additional Experimental Results}

\noindent\textbf{Detailed Grounding DINO Results on COCO-C.}
\Cref{tab:grounding_dino_detailed} reports the corruption-wise results corresponding to the grouped comparison presented in the main paper. DESA-TTA improves over both ZS inference and TTAOD-F across all 15 corruption types and achieves the best performance on 9 corruptions. Relative to the strongest baseline for each corruption, the largest improvements are obtained on contrast ($+5.1$ mAP), glass blur ($+2.4$), impulse noise ($+1.8$), and snow ($+1.6$). Overall, DESA-TTA achieves 27.6 mAP, outperforming ZS inference, TTAOD-F, and VLOD-TTA by 7.0, 1.6, and 1.4 points, respectively.\\

\begin{table*}[t]
\centering
\caption{\textbf{Detailed Grounding DINO results on COCO-C.} mAP is reported for each corruption type, with the best result in each column highlighted in bold.}
\label{tab:grounding_dino_detailed}
\resizebox{\textwidth}{!}{
\begin{tabular}{lccccccccccccccc|c}
\toprule
\multirow{2}{*}{\textbf{Method}}
& \multicolumn{3}{c}{\textbf{Noise}}
& \multicolumn{4}{c}{\textbf{Blur}}
& \multicolumn{3}{c}{\textbf{Weather}}
& \multicolumn{5}{c}{\textbf{Digital}}
& \\
\cmidrule(lr){2-4}
\cmidrule(lr){5-8}
\cmidrule(lr){9-11}
\cmidrule(lr){12-16}
& Gau. & Shot & Imp.
& Def. & Glass & Mot. & Zoom
& Snow & Frost & Fog
& Bright. & Cont. & Elast. & Pix. & JPEG
& Avg. \\
\midrule
ZS
& 13.7 & 16.0 & 15.0
& 16.8 & 7.5 & 13.6 & 6.7
& 27.5 & 32.5 & 44.2
& 44.1 & 21.9 & 22.5 & 5.3 & 21.1
& 20.6 \\

TTAOD-F
& 20.2 & 22.0 & 21.4
& 17.8 & 14.5 & 16.9 & 7.9
& 31.1 & 34.7 & 45.1
& 44.9 & 30.6 & 29.9 & 23.6 & 29.2
& 26.0 \\

VLOD-TTA
& 20.5 & 21.8 & 20.9
& \textbf{20.6} & 13.1 & \textbf{18.4} & \textbf{10.1}
& 30.8 & \textbf{36.6} & \textbf{48.2}
& \textbf{46.2} & 28.8 & 28.7 & 19.7 & 28.6
& 26.2 \\
\midrule
\textbf{DESA-TTA}
& \textbf{21.9} & \textbf{22.8} & \textbf{23.2}
& 20.4 & \textbf{16.9} & 17.6 & 8.5
& \textbf{32.7} & 36.1 & 47.9
& 45.4 & \textbf{35.7} & \textbf{30.1} & \textbf{24.5} & \textbf{29.9}
& \textbf{27.6} \\
\bottomrule
\end{tabular}
}
\end{table*}

\noindent\textbf{Generalization Across Backbones. }
To assess whether DESA-TTA remains effective for a larger VLOD, we evaluate it using YOLO-World-Large on the style-shift benchmarks. As shown in Table~\ref{tab:yoloworld_large}, the same trend observed with YOLO-World-Small is preserved at a larger model scale. Standard mean-teacher adaptation improves over zero-shot inference, while DESA-TTA further improves performance across all three datasets. On average, DESA-TTA improves AP$_{50}$ by $+8.4$ points over zero-shot inference, $+4.6$ points over VLOD-TTA, and $+3.2$ points over standard mean-teacher adaptation. These results indicate that DESA-TTA remains effective beyond the YOLO-World-S setting.\\

\begin{table}[!ht]
\centering
\caption{\textbf{Performance with YOLO-World-L on style-shift benchmarks.} AP$_{50}$ is reported.}
\label{tab:yoloworld_large}
\resizebox{\linewidth}{!}{
\begin{tabular}{lccc|c}
\toprule
\textbf{Method} & \textbf{Watercolor} & \textbf{Comic} & \textbf{ClipArt} & \textbf{Avg.} \\
\midrule
ZS       & 55.3 & 37.9 & 50.6 & 47.9 \\
VLOD-TTA & 58.3 & 42.8 & 53.9 & 51.7 \\
MT       & 58.9 & 45.1 & 55.3 & 53.1 \\
\midrule
\textbf{DESA-TTA} & \textbf{61.8} & \textbf{48.3} & \textbf{58.7} & \textbf{56.3} \\
\bottomrule
\end{tabular}
}
\end{table}

\noindent\textbf{Comparison with CoTTA. }
We further compare DESA-TTA with CoTTA~\citep{cotta}, a continual TTA method that mitigates forgetting by randomly restoring a subset of model parameters to their source-pretrained values. As shown in~\Cref{tab:cotta_comparison}, CoTTA improves over standard MT on VOC-C, indicating that source restoration can help under long corruption streams. However, random restoration remains less effective on the shorter style-shift streams. Adding dynamic EMA to CoTTA improves both Style and VOC-C, but DESA-TTA achieves the best performance on both benchmarks. These results show that random source restoration can improve robustness, while student-drift-dependent source anchoring provides more effective control of cumulative student drift.\\

\begin{table}[t]
\centering
\caption{\textbf{Comparison with source-restoration baselines.} Average AP$_{50}$ is reported on Style and VOC-C benchmarks.}
\label{tab:cotta_comparison}
\resizebox{.30\textwidth}{!}{
\begin{tabular}{lcc}
\toprule
Method & Style & VOC-C \\
\midrule
ZS              & 39.1 & 34.0 \\
MT              & 44.1 & 39.5 \\
CoTTA           & 44.6 & 44.4 \\
D-EMA + CoTTA   & 46.0 & 47.1 \\
\midrule
\textbf{DESA-TTA} & \textbf{47.3} & \textbf{48.5} \\
\bottomrule
\end{tabular}
}
\vspace{-1mm}
\end{table}

\noindent\textbf{Comparison with $L_2$ Regularization. }
To determine whether conventional regularization can effectively control student drift, we compare source anchoring with a source-centered $L_2$ baseline that augments the detection loss with
$\lambda_{\mathrm{reg}}\sum_{\ell\in\mathcal{A}}
\|\theta_{\ell}^{S}-\theta_{0,\ell}\|_2^2$.
Unlike source anchoring, which interpolates the gradient-updated student parameters with their pretrained values, this baseline imposes the source constraint directly through the optimization objective. All variants use dynamic temporal averaging, thereby isolating the effect of the student constraint. As shown in~\Cref{tab:source_regularization}, $L_2$ regularization improves VOC-C performance from 43.1 to 44.2 AP$_{50}$ but reduces Style performance from 46.6 to 45.1. Static source anchoring provides a larger improvement on VOC-C, reaching 45.8 AP$_{50}$, but also remains below D-EMA on Style. Thus, neither penalty-based regularization nor static source anchoring provides consistent improvements across both benchmarks. In contrast, drift-dependent source anchoring achieves the best performance on both Style and VOC-C, outperforming $L_2$ regularization by 2.2 and 4.3 AP$_{50}$, respectively.\\

\begin{table}[t]
\centering
\caption{\textbf{Comparison of source anchoring and $L_2$ regularization.} Average AP$_{50}$ is reported on Style and VOC-C benchmarks.}
\label{tab:source_regularization}
\begin{tabular}{lcc}
\toprule
Method & Style & VOC-C \\
\midrule
D-EMA                         & 46.6 & 43.1 \\
D-EMA + $L_2$ regularization & 45.1 & 44.2 \\
D-EMA + SA                    & 46.4 & 45.8 \\
\midrule
\textbf{DESA-TTA}                      & \textbf{47.3} & \textbf{48.5} \\
\bottomrule
\end{tabular}
\end{table}

\noindent\textbf{Variation Across Random Seeds. }
All results reported in the main paper are averaged over three independent runs with different random seeds for each dataset. \Cref{tab:std_results} reports the corresponding standard deviations. Across both benchmark groups, the standard deviation does not exceed 0.24 AP$_{50}$ for any method, while DESA-TTA retains the highest mean performance, indicating consistent performance across runs.\\

\begin{table}[t]
\centering
\caption{\textbf{Mean AP$_{50}$ $\pm$ standard deviation over three seeds.} Style is averaged over Watercolor, ClipArt, and Comic, and VOC-C over all 15 corruptions.}
\label{tab:std_results}
\begin{tabular}{lcc}
\toprule
Method & Style & VOC-C \\
\midrule
ZS       &  39.1           &  34.0          \\
MT       & 44.1 $\pm$ 0.21 & 39.5 $\pm$ 0.16 \\
D-EMA    & 46.6 $\pm$ 0.13 & 43.1 $\pm$ 0.21 \\
\midrule
\textbf{DESA-TTA} & \textbf{47.3 $\pm$ 0.24} & \textbf{48.5 $\pm$ 0.15} \\
\bottomrule
\vspace{-.1cm}
\end{tabular}
\end{table}

\begin{figure}[t]
    \centering
    \includegraphics[width=\linewidth]{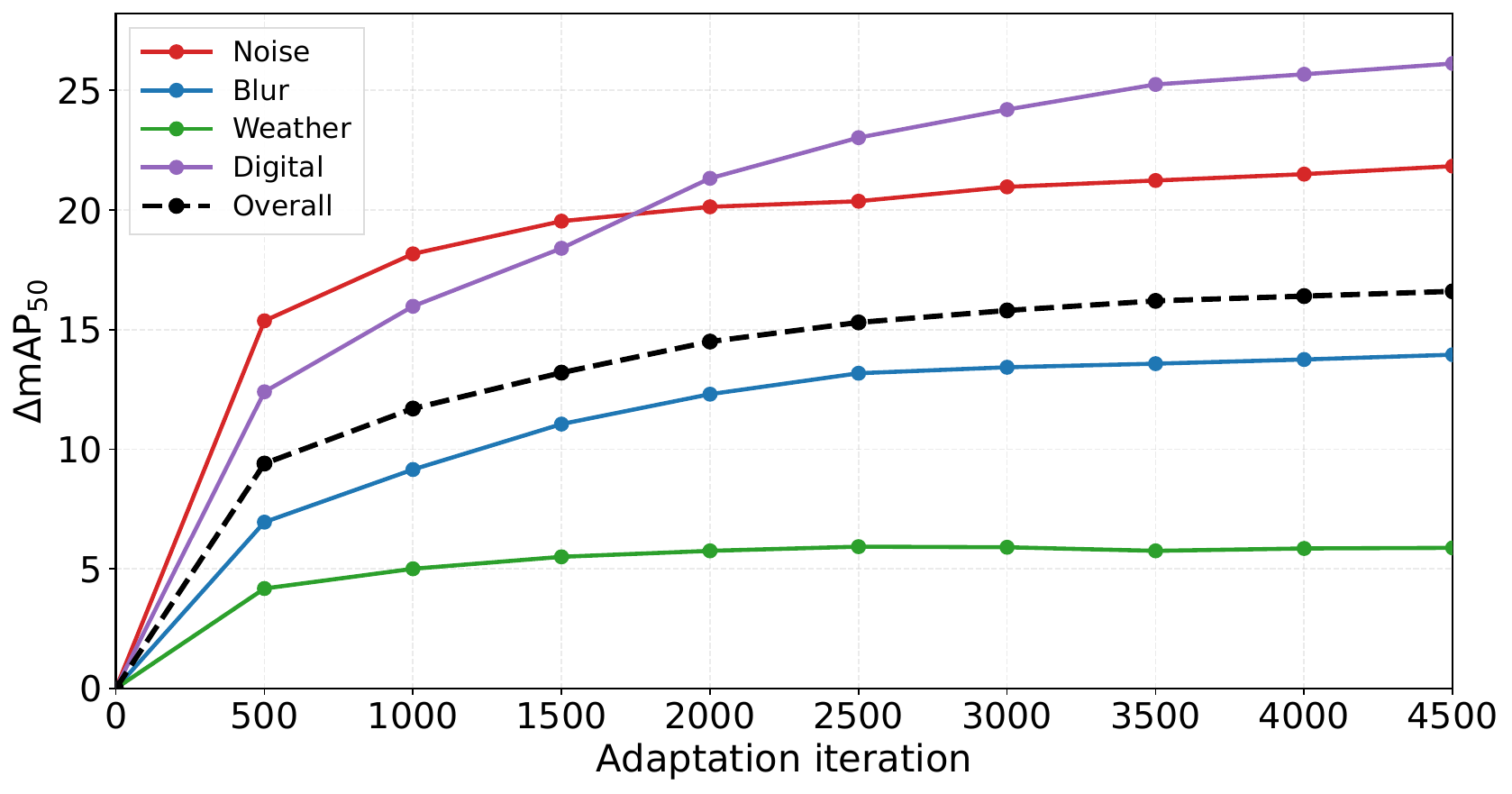}
    \caption{\textbf{Generalization to a held-out subset of VOC-C.}
    DESA-TTA snapshots are evaluated at different adaptation iterations on held-out target data. We report AP$_{50}$ gains over the ZS checkpoint for each corruption group and the overall average.}
    \label{fig:generalization_ability}
    \vspace{-4mm}
\end{figure}

\noindent\textbf{Generalization to Held-Out Target Data. }
DESA-TTA performs cumulative adaptation by continuously updating the model over the target stream. To assess whether the adapted model generalizes beyond the samples used for adaptation, we hold out VOC-C target images from the adaptation stream and evaluate model snapshots saved at different adaptation iterations on this held-out subset. As shown in~\Cref{fig:generalization_ability}, AP$_{50}$ consistently increases with the number of adaptation iterations, yielding an average gain of $+16.6$ AP$_{50}$ over the ZS model. The adaptation process takes approximately $3$ minutes on a single RTX 3090 GPU. These results indicate a practical deployment setting in which the detector can be adapted to a specific target environment using an unlabeled target stream and then deployed at full inference speed.

\begin{figure*}[t]
    \centering

    \begin{minipage}[t]{0.33\textwidth}
        \centering
        \includegraphics[width=\linewidth]{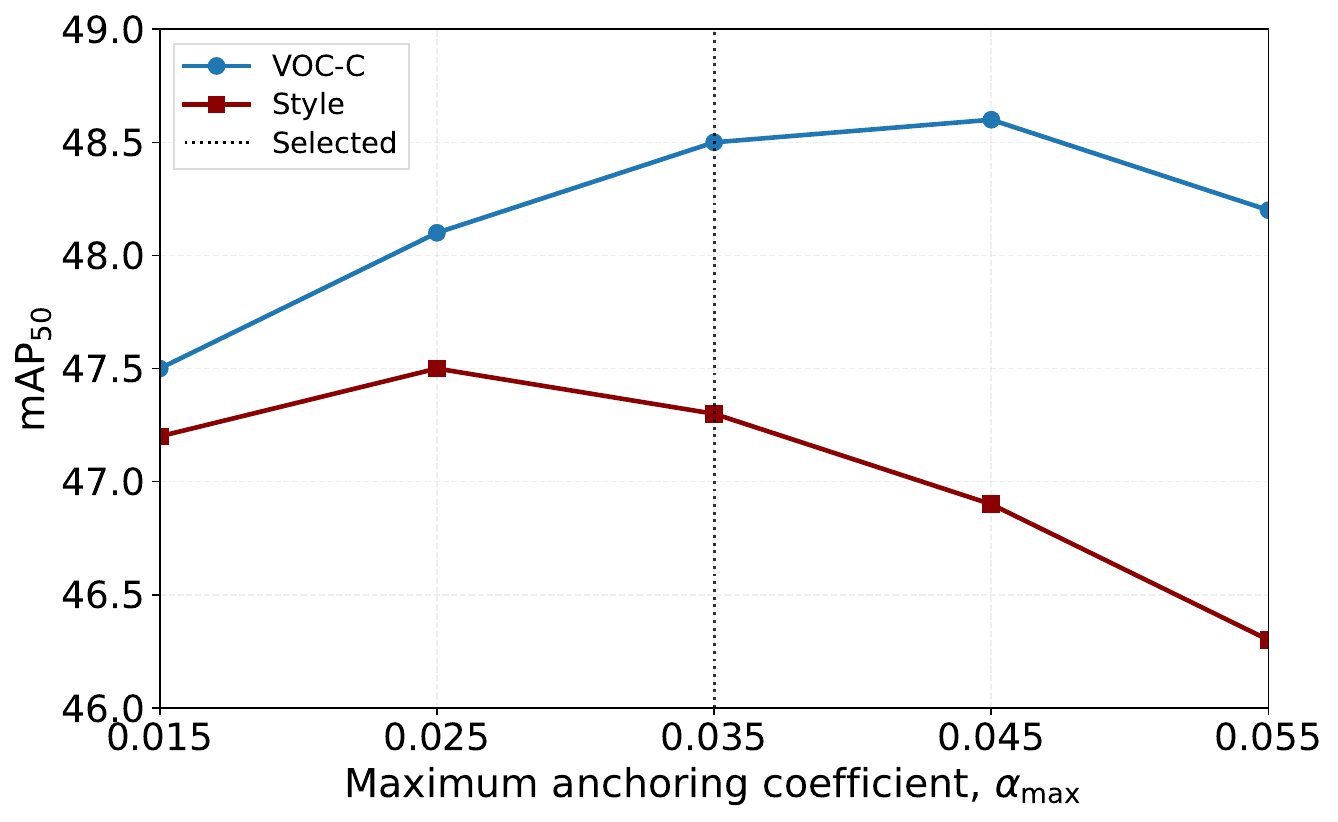}
        \caption*{(a) Maximum anchoring coefficient.}
    \end{minipage}
    \hfill
    \begin{minipage}[t]{0.33\textwidth}
        \centering
        \includegraphics[width=\linewidth]{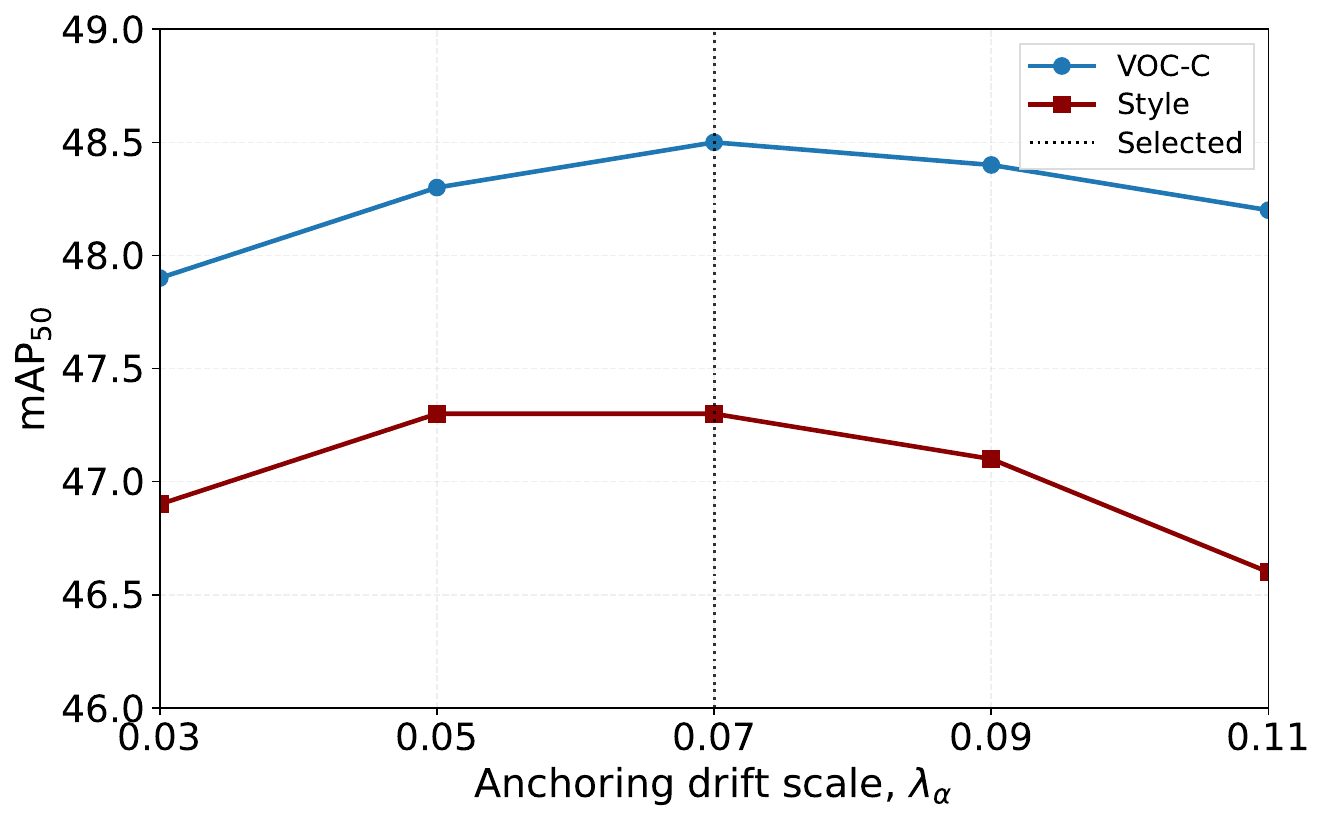}
        \caption*{(b) Anchoring drift scale.}
    \end{minipage}
    \hfill
    \begin{minipage}[t]{0.33\textwidth}
        \centering
        \includegraphics[width=\linewidth]{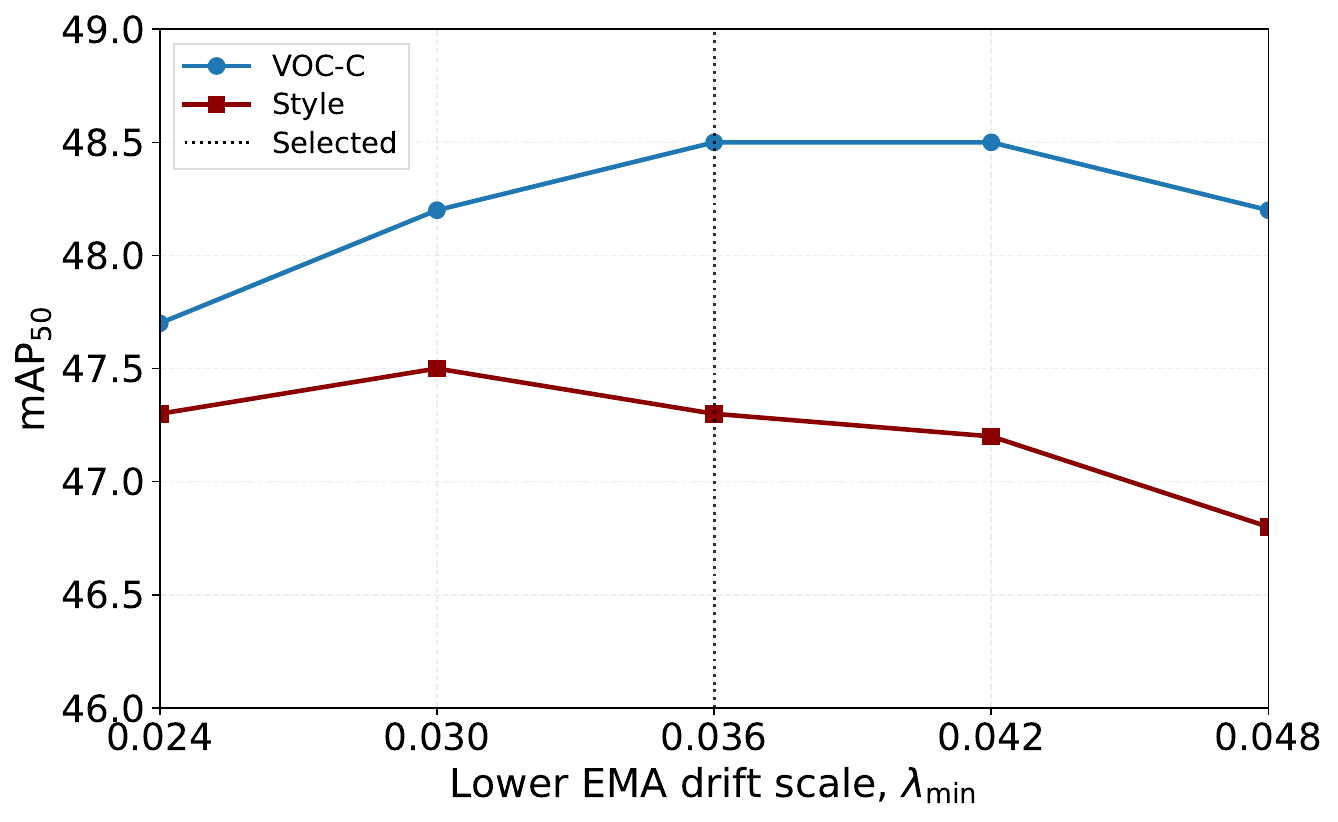}
        \caption*{(c) Lower EMA drift scale.}
    \end{minipage}

    \par\vspace{2mm}

    \begin{minipage}[t]{0.33\textwidth}
        \centering
        \includegraphics[width=\linewidth]{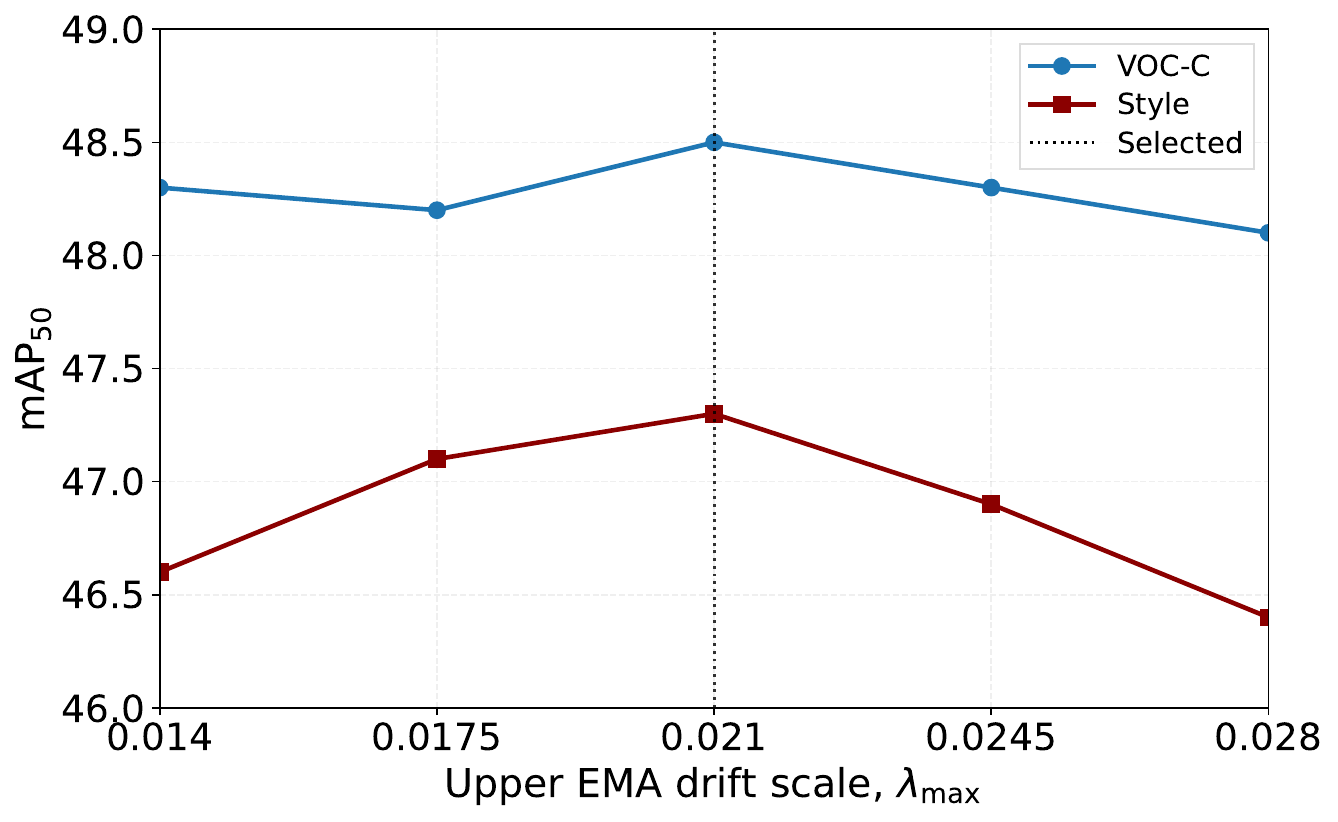}
        \caption*{(d) Upper EMA drift scale.}
    \end{minipage}
    \hspace{0.04\textwidth}
    \begin{minipage}[t]{0.33\textwidth}
        \centering
        \includegraphics[width=\linewidth]{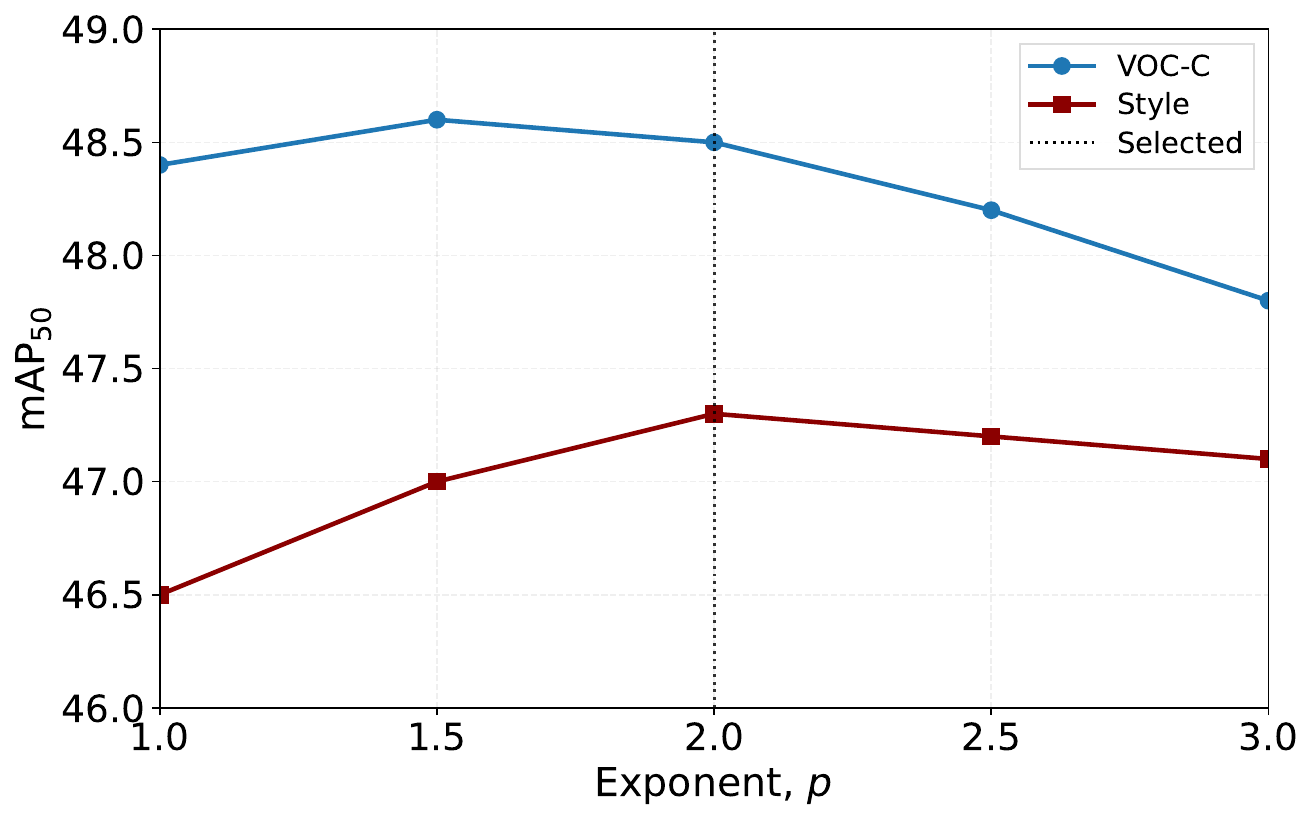}
        \caption*{(e) Transition exponent.}
    \end{minipage}

    \caption{\textbf{Hyperparameter sensitivity of DESA-TTA on VOC-C and the style-shift datasets.}
    Average AP$_{50}$ is reported, with Style averaged over Watercolor, ClipArt, and Comic and VOC-C averaged over the 15 corruption types. Each hyperparameter is varied independently while the remaining settings are fixed at their selected values. The vertical dotted line indicates the selected value.}
    \label{fig:hyperparameter_analysis}
\end{figure*}

\subsection{Hyperparameter Analysis}
The hyperparameter values used in all experiments were selected using six corruption types from the VOC-C training set, as described in the main paper. \Cref{fig:hyperparameter_analysis} provides a sensitivity analysis of five hyperparameters: $\alpha_{\max}$ and $\lambda_{\alpha}$ for source anchoring, $\lambda_{\min}$ and $\lambda_{\max}$ for dynamic temporal averaging, and the exponent $p$, which is used in both components. These parameters control the maximum anchoring strength and the responses of both components to parameter drift. We do not treat $M_{\max}$ as a tunable hyperparameter because it is fixed to the detector's maximum number of retained predictions. We fix $m_{\min}$, $m_{\max}$, $\Delta_{\min}$, and $\Delta_{\max}$ because they jointly define the low- and high-drift EMA intervals, chosen based on the effective coefficient range observed in the fixed-EMA analysis. Each analyzed hyperparameter is varied over five values while the others are fixed at their selected values.\\

\noindent\textbf{Maximum Anchoring Coefficient.}
\Cref{fig:hyperparameter_analysis}(a) shows different responses to the anchoring strength across the two benchmark groups. VOC-C performance improves as $\alpha_{\max}$ increases up to $0.045$, suggesting that stronger restoration toward the pretrained model is beneficial over long corruption streams, where repeated pseudo-label updates create a greater risk of cumulative student drift. In contrast, Style performance peaks at $\alpha_{\max}=0.025$ and subsequently decreases, indicating that excessive anchoring can restrict the parameter changes required to adapt to substantial appearance shifts. Performance on both benchmarks decreases at the largest value, showing that overly strong anchoring eventually suppresses useful target adaptation. Our setting, $\alpha_{\max}=0.035$, lies between the two benchmark-specific optima and achieves 47.3 AP$_{50}$ on Style and 48.5 AP$_{50}$ on VOC-C.\\

\noindent\textbf{Anchoring Drift Scale.}
\Cref{fig:hyperparameter_analysis}(b) examines the anchoring drift scale $\lambda_{\alpha}$, which controls how rapidly the anchoring coefficient increases with student drift. When $\lambda_{\alpha}$ is small, the anchoring coefficient becomes large at relatively low drift, pulling the updated student parameters strongly toward their pretrained values before sufficient adaptation has occurred. When $\lambda_{\alpha}$ is too large, anchoring remains weak even as student drift increases, reducing its ability to limit accumulated pseudo-label errors. The selected value $\lambda_{\alpha}=0.07$ lies in this intermediate regime, achieving the highest VOC-C performance of 48.5 AP$_{50}$ and tying the highest Style performance of 47.3 AP$_{50}$.\\

\noindent\textbf{Lower EMA Drift Scale.}
\Cref{fig:hyperparameter_analysis}(c) evaluates the lower EMA drift scale $\lambda_{\min}$, which controls how quickly the lower EMA bound increases from its reduced initial value as teacher drift grows. Increasing $\lambda_{\min}$ keeps the lower bound reduced over a wider drift range, allowing the teacher to incorporate student updates more rapidly for longer during adaptation. This improves VOC-C performance up to $\lambda_{\min}=0.036$, whereas Style performance begins to decrease beyond $0.030$, suggesting that maintaining a lower EMA bound for too long can make the teacher overly responsive to recent student updates on shorter target streams. We select $\lambda_{\min}=0.036$, which attains the highest VOC-C performance of 48.5 AP$_{50}$ while remaining within 0.2 points of the best Style result.\\

\noindent\textbf{Upper EMA Drift Scale.}
\Cref{fig:hyperparameter_analysis}(d) examines the upper EMA drift scale $\lambda_{\max}$, which controls how quickly the upper EMA bound increases from its reduced initial value as teacher drift grows. When $\lambda_{\max}$ is small, the upper bound increases rapidly, imposing strong temporal averaging before the teacher has sufficiently adapted to the target distribution. Conversely, a large $\lambda_{\max}$ keeps the upper bound reduced over a wider drift range, weakening temporal smoothing and increasing the teacher's sensitivity to recent student updates. At $\lambda_{\max}=0.021$, both benchmark groups attain their highest performance, reaching 47.3 AP$_{50}$ on Style and 48.5 AP$_{50}$ on VOC-C.\\

\noindent\textbf{Transition Exponent.}
Finally, \Cref{fig:hyperparameter_analysis}(e) examines the exponent $p$, which controls the sharpness of the drift-dependent transition in the source-anchoring coefficient and both EMA bounds. A small value produces gradual transitions that weakly differentiate between low- and high-drift states, whereas a large value makes the transitions increasingly abrupt and therefore more sensitive to small changes around their respective drift scales. The results favor an intermediate transition: Style performance peaks at $p=2$, while VOC-C peaks at $p=1.5$ and differs by only 0.1 AP$_{50}$ at $p=2$. At the selected value $p=2$, DESA-TTA achieves 47.3 AP$_{50}$ on Style and 48.5 AP$_{50}$ on VOC-C. The resulting configuration uses $\alpha_{\max}=0.035$, $\lambda_{\alpha}=0.07$, $\lambda_{\min}=0.036$, $\lambda_{\max}=0.021$, and $p=2$.

\subsection{Qualitative Analysis}
\begin{figure*}[t]
    \centering
    \setlength{\tabcolsep}{0pt}

    \begin{tabular}{
        @{}
        *{5}{>{\centering\arraybackslash}p{0.2\textwidth}}
        @{}
    }
        \textbf{GT} &
        \textbf{ZS} &
        \textbf{MT} &
        \textbf{Dynamic EMA} &
        \textbf{DESA-TTA}
        \\[1pt]

        \multicolumn{5}{c}{
            \includegraphics[width=\textwidth]{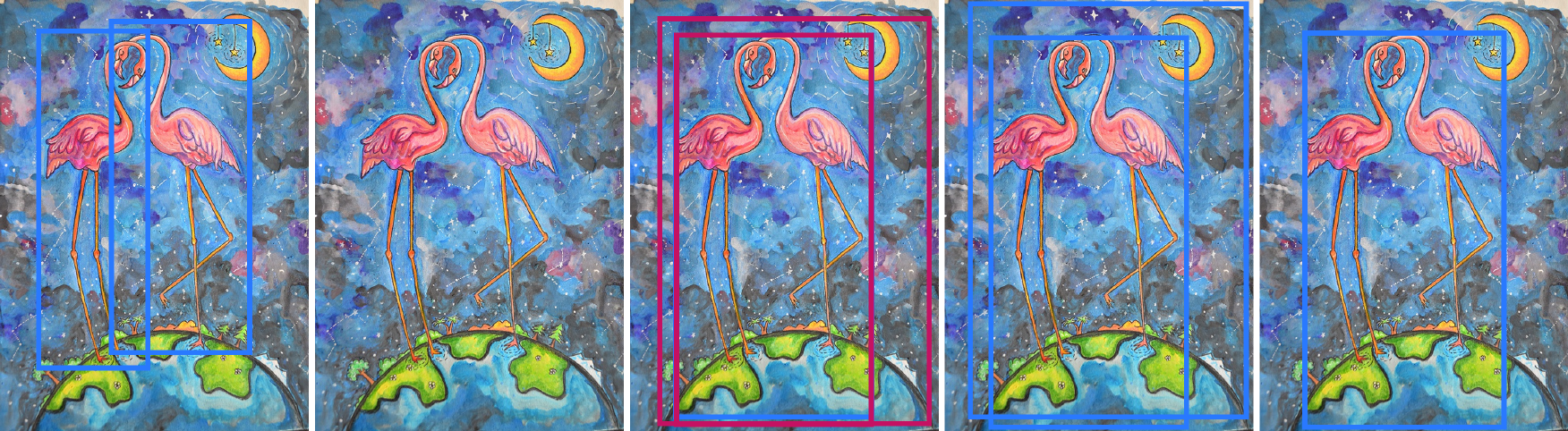}
        }
        \\[.1pt]

        \multicolumn{5}{c}{
            \includegraphics[width=\textwidth]{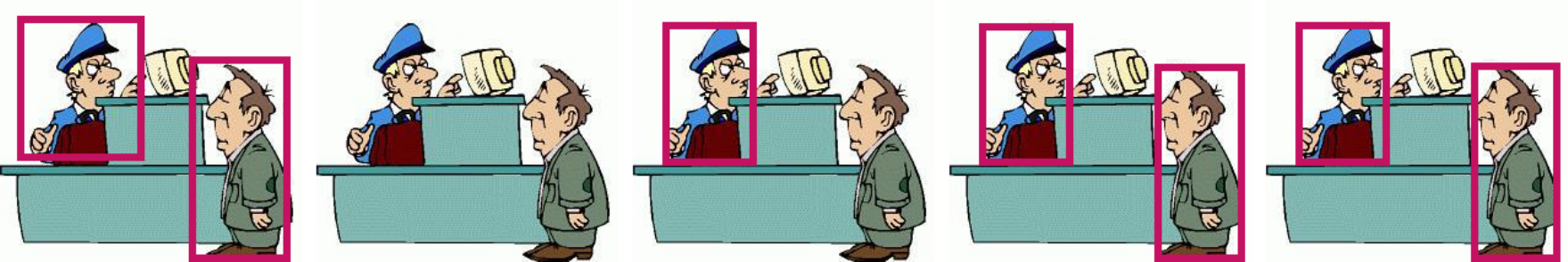}
        }
        \\[.1pt]

        \multicolumn{5}{c}{
            \includegraphics[width=\textwidth]{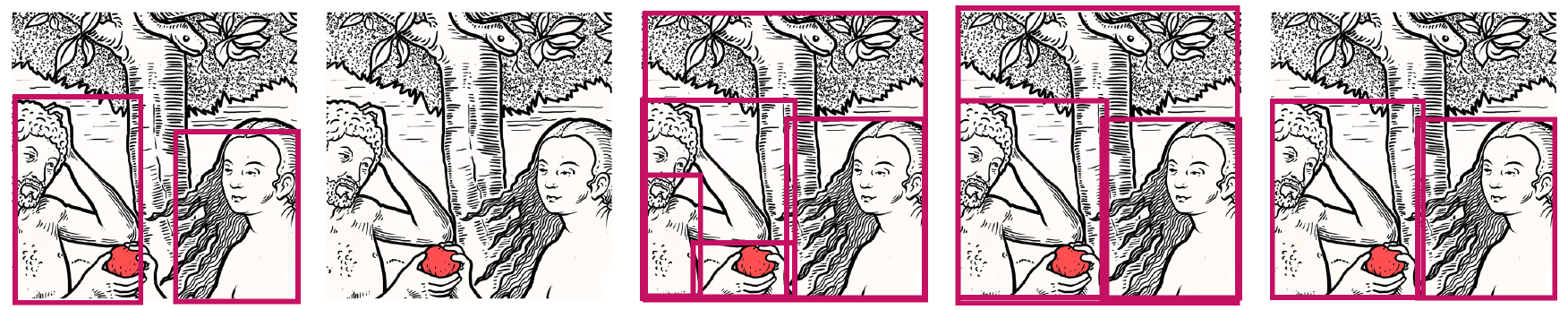}
        }
        \\[.1pt]


        \multicolumn{5}{c}{
            \includegraphics[width=\textwidth]{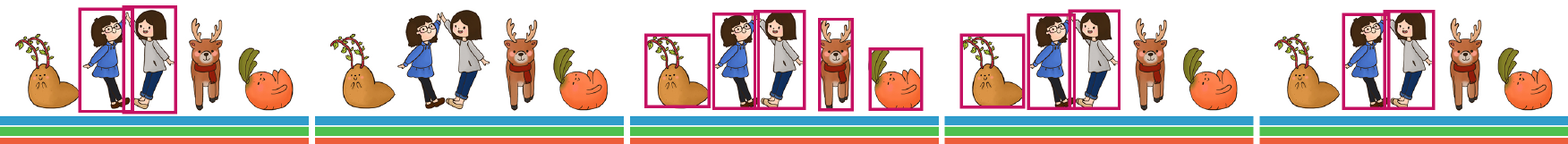}
        }
    \end{tabular}

\caption{\textbf{Qualitative comparison under style shifts.}
The first two rows show examples from Watercolor and ClipArt, respectively, while the last two rows show Comic examples. Columns compare ground truth (GT), zero-shot inference (ZS), standard mean-teacher adaptation (MT), dynamic temporal averaging without source anchoring (Dynamic EMA), and DESA-TTA.}
\label{fig:qualitative_style}
    
\end{figure*}

\noindent\textbf{Style Shift. }
\Cref{fig:qualitative_style} presents qualitative comparisons on Watercolor, ClipArt, and Comic. In the Watercolor example, ZS produces no detections for the two annotated birds, whereas DESA-TTA identifies both instances. In the ClipArt example, ZS misses both annotated persons and MT recovers only one, while Dynamic EMA and DESA-TTA detect both. In the Comic examples, MT introduces several detections that do not correspond to annotated objects. Dynamic EMA reduces these false positives, while DESA-TTA further suppresses the remaining spurious detections without removing the correctly detected persons. These examples indicate that DESA-TTA can improve object recovery while limiting false-positive predictions under style shifts.\\

\begin{figure*}[t]
    \centering
    \setlength{\tabcolsep}{0pt}

    \begin{tabular}{
        @{}
        *{5}{>{\centering\arraybackslash}p{0.2\textwidth}}
        @{}
    }
        \textbf{GT} &
        \textbf{ZS} &
        \textbf{MT} &
        \textbf{Dynamic EMA} &
        \textbf{DESA-TTA}
        \\[1pt]

        \multicolumn{5}{c}{
            \includegraphics[width=\textwidth]{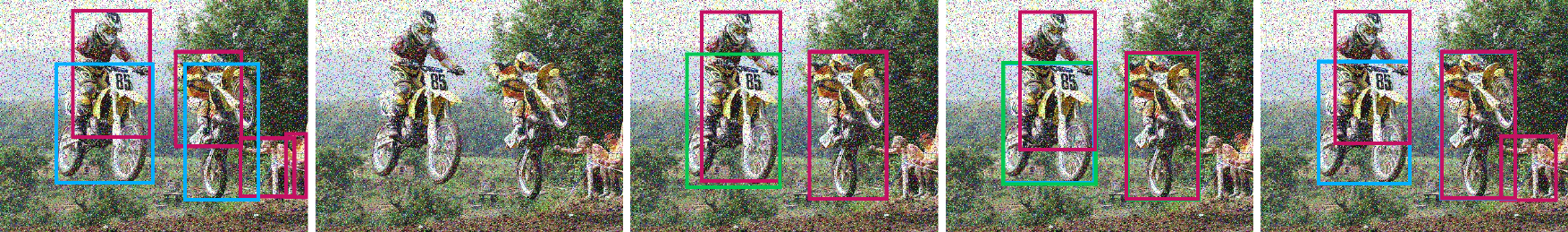}
        }
        \\[.1pt]

        \multicolumn{5}{c}{
            \includegraphics[width=\textwidth]{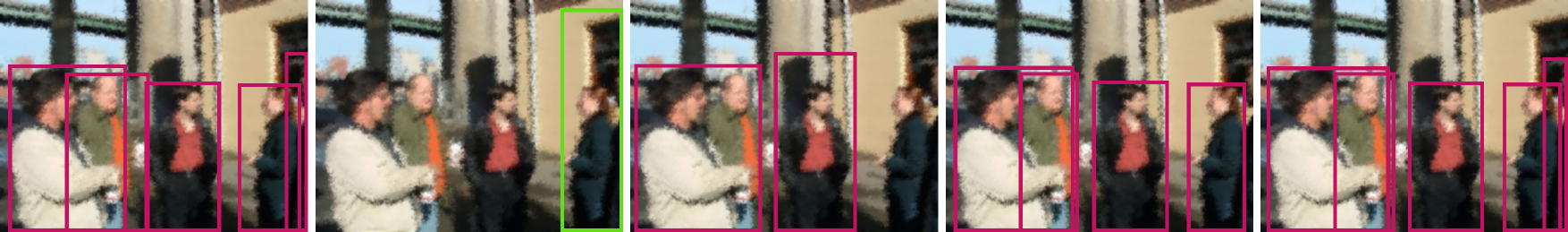}
        }
        \\[.1pt]

        \multicolumn{5}{c}{
            \includegraphics[width=\textwidth]{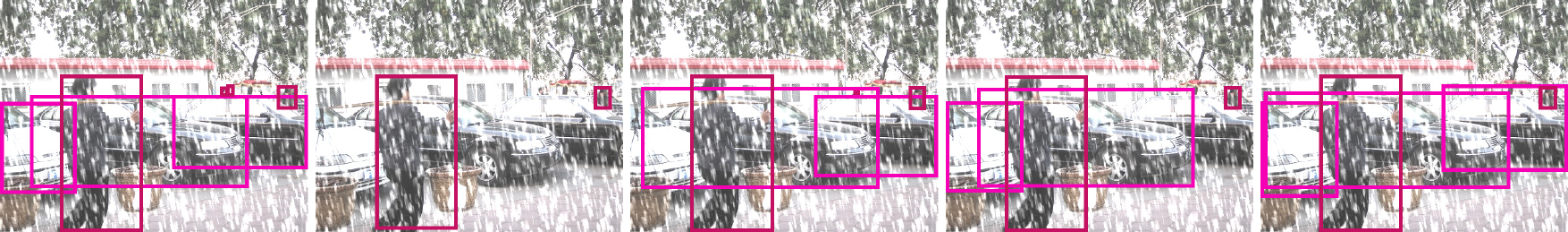}
        }
        \\[.1pt]

        \multicolumn{5}{c}{
            \includegraphics[width=\textwidth]{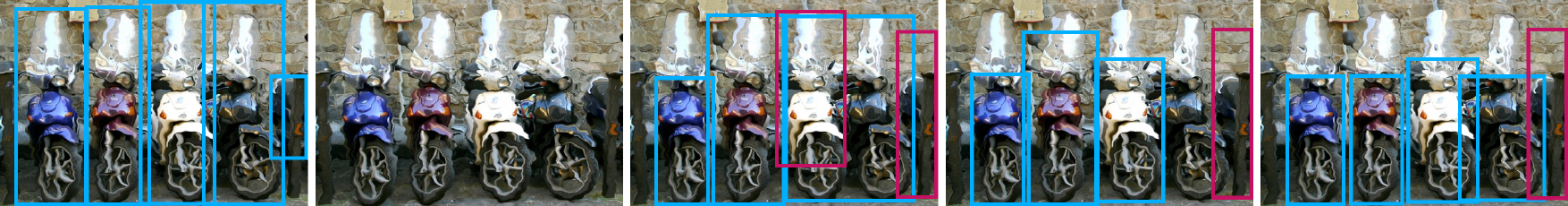}
        }

    \end{tabular}

\caption{\textbf{Qualitative comparison under common corruptions.}
Rows show examples under impulse noise, glass blur, snow, and elastic transformation, from top to bottom. Columns compare ground truth (GT), zero-shot inference (ZS), standard mean-teacher adaptation (MT), dynamic temporal averaging without source anchoring (Dynamic EMA), and DESA-TTA.}
\label{fig:qualitative_corruptions}
\end{figure*}

\noindent\textbf{Common Corruptions. } 
\Cref{fig:qualitative_corruptions} presents qualitative comparisons under impulse noise, glass blur, snow, and elastic transformation. Under impulse noise, ZS fails to detect the annotated objects, while MT and Dynamic EMA identify the two prominent riders but miss the smaller instance on the right. DESA-TTA additionally identifies this instance while retaining the other detections. Under glass blur, DESA-TTA recovers a more complete set of annotated persons than the compared variants. Under snow, it preserves detections for several objects missed by ZS, and under elastic transformation, it identifies more of the densely arranged motorcycles. These observations are consistent with the quantitative results, indicating that DESA-TTA improves instance recovery across diverse corruption types.